\documentclass[default]{sn-jnl}

\usepackage{graphicx}%
\usepackage{multirow}%
\usepackage{amsmath,amssymb,amsfonts}%
\usepackage[title]{appendix}%
\usepackage{xcolor}%
\usepackage{manyfoot}%
\usepackage{booktabs}%
\usepackage{listings}%
\usepackage{subcaption}
\usepackage{adjustbox}

\usepackage{pifont}
\newcommand{\cmark}{\ding{51}}
\newcommand{\xmark}{-}%
\usepackage[titlenumbered, ruled, linesnumbered, noend]{algorithm2e}
\SetKwComment{tcp}{$\rhd$ }{}

\SetCommentSty{mycommfont}
\SetKw{And}{and}
\SetKw{Break}{break}
\SetKw{Or}{or}
\SetKw{In}{in}
\SetKw{Continue}{continue}
\let\oldnl\nl
\newcommand{\nonl}{\renewcommand{\nl}{\let\nl\oldnl}}
\usepackage{setspace}
\usepackage{orcidlink}
\usepackage{enumitem}
\usepackage{natbib}
\setcitestyle{authoryear}
\setcitestyle{round}
\setcitestyle{semicolon}
\setcitestyle{aysep={}}
\usepackage[textsize=scriptsize]{todonotes}
\usepackage{marginnote}

\usepackage{cancel}
\usepackage[normalem]{ulem}
\usepackage{soul}
\usepackage{xcolor}

\begin{document}

\title[Article Title]{LoCoMotif: Discovering time-warped motifs in time series}

\author*[1,2]{\fnm{Daan} \sur{Van Wesenbeeck} \orcidlink{0000-0002-4941-5480}}\email{daan.vanwesenbeeck@kuleuven.be}
\author[1,2]{\fnm{Aras} \sur{Yurtman} \orcidlink{0000-0001-6213-5427}}
\author[1,2]{\fnm{Wannes} \sur{Meert} \orcidlink{0000-0001-9560-3872}}
\author[1,2]{\fnm{Hendrik} \sur{Blockeel} \orcidlink{0000-0003-0378-3699}}

\affil[1]{\orgdiv{Dept. of Computer Science}, \orgname{KU Leuven}, \orgaddress{\city{Leuven}, \postcode{B-3000}, \country{Belgium}}}
\affil[2]{\orgdiv{Leuven.AI - KU Leuven Institute for AI}, \city{Leuven}, \postcode{B-3000}, \country{Belgium}}
\abstract{
Time Series Motif Discovery (TSMD) refers to the task of identifying patterns that occur multiple times (possibly with minor variations) in a time series. 
All existing methods for TSMD have one or more of the following limitations: they only look for the two most similar occurrences of a pattern;
they only look for patterns of a pre-specified, fixed length; they cannot handle variability along the time axis; and they only handle univariate time series. 
In this paper, we present a new method, LoCoMotif, that has none of these limitations. 
The method is motivated by a concrete use case from physiotherapy. We demonstrate the value of the proposed method on this use case. We also introduce a new quantitative evaluation metric for motif discovery, and benchmark data for comparing TSMD methods. LoCoMotif substantially outperforms the existing methods, on top of being more broadly applicable.
}


\maketitle

\section{Introduction}\label{sec:introduction}

Time Series Motif Discovery (TSMD) is defined as the task of finding patterns that approximately repeat in a given time series. TSMD finds application in a wide variety of domains, such as cardiology \citep{ecg_mp}, physiotherapy \citep{Lin2016}, audio (speech) processing \citep{motiflets, hao_parameter-free_2013}, seismology \citep{zimmerman_matrix_2019}, and more. Additionally, TSMD is used in other time series tasks such as anomaly detection, clustering and classification \citep{motiflets}.

As the terminology used in the literature varies somewhat, it is useful to precisely define some terms before continuing.  In this paper, the term {\em motif} refers to a single occurrence of a pattern.  A {\em motif set} is a set of occurrences that are all similar enough to be considered the same pattern, and a {\em motif pair} is a motif set with cardinality two.

All existing methods for TSMD have a number of limitations that make them inapplicable in certain contexts. First, many methods find only motif pairs, rather than motif sets of any cardinality. Second, many methods can only find motifs of a pre-specified length (so all motifs within and across motif sets must have identical length).  Third, similarity is typically measured using Euclidean distance, even though in many contexts an elastic measure such as Dynamic Time Warping (DTW) distance is considered more relevant. Finally, most TSMD methods assume univariate time series, while many practical contexts feature multivariate time series. 

\begin{figure}
\begin{subfigure}[t]{\linewidth}
\includegraphics[width=\linewidth]{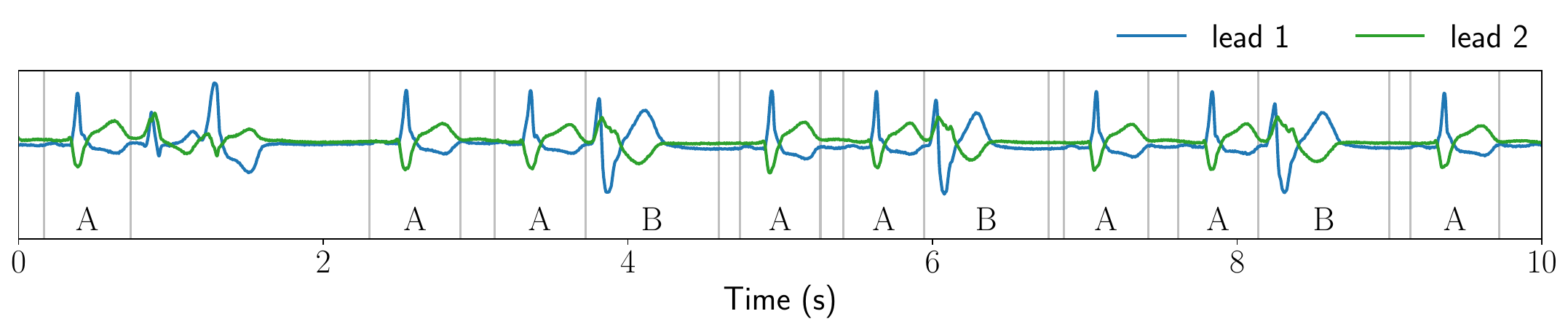}
\caption{}
\label{fig:ecg_example}
\end{subfigure}\hfill
\begin{subfigure}[t]{0.3497\linewidth}
\includegraphics[height=3.2cm]{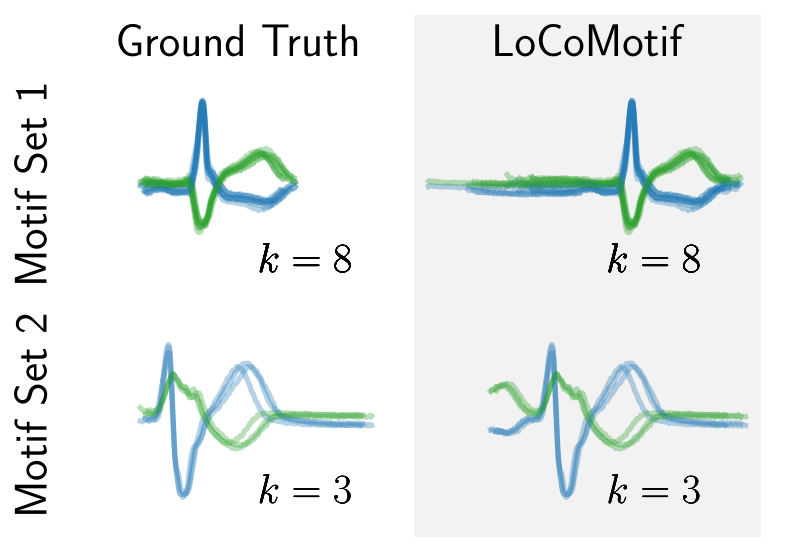}
\caption{}
\label{fig:ecg_gt_locomotif}
\end{subfigure}
\hfill
\begin{subfigure}[t]{0.63029\linewidth}
\centering
\includegraphics[height=3.2cm]{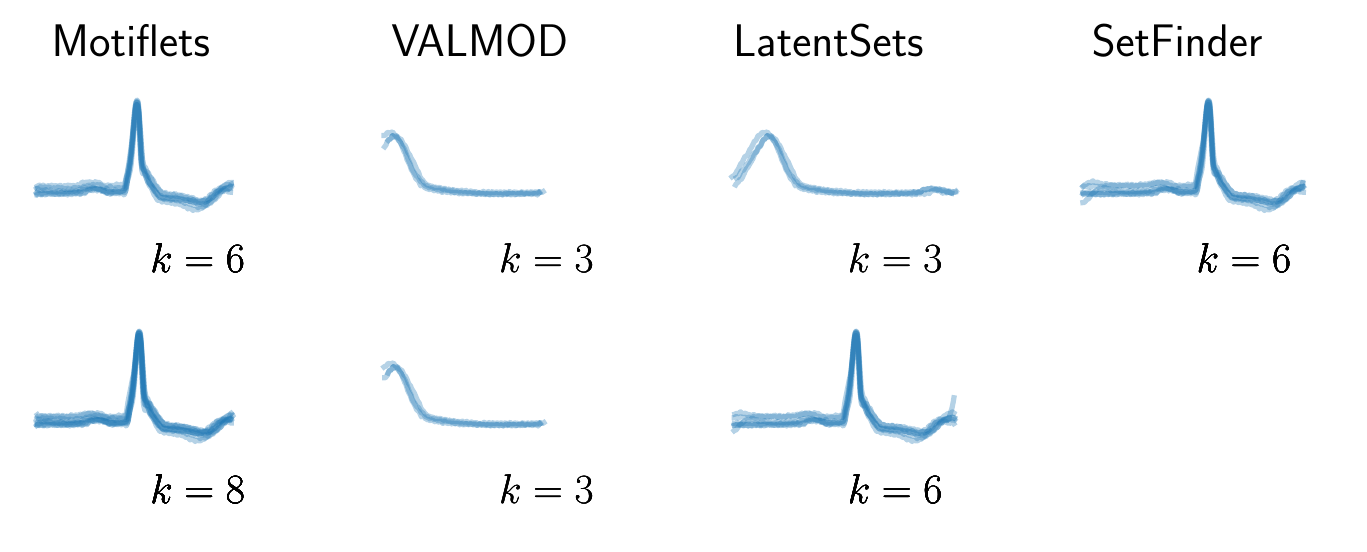}
\caption{}
\label{fig:ecg_existing}
\end{subfigure}
\caption{In an ECG example (a) that contains two types of heartbeats (type A and B for simplicity), LoCoMotif discovers both types of heartbeats (b), whereas existing methods cannot (c). Since existing methods are not applicable to multivariate time series, they are applied only on lead 1. $k$ denotes the cardinality of a motif set.}
\label{fig:ecg}
\end{figure}

Due to these limitations, there are application domains where TSMD would be useful but none of the existing methods can be applied.  As an example, consider electrocardiograms (ECGs). Fig.~\ref{fig:ecg_example} shows a so-called two-lead ECG; it consists of two parallel signals (leads).\footnote{The example is part of the recording of patient 214 in the MIT-BIH Arrhythmia Database \citep{mitbih}.} Two different types of heartbeats occur in this ECG; each is a separate motif set. The TSMD task consists of recognizing that there are two different types of heartbeats, and identifying their occurrences.  We note that there are more than two occurrences of each type of heartbeat; the two types differ in length (0.6 and 0.9 seconds); there is small temporal variation within the second type (visible in Fig.~\ref{fig:ecg_gt_locomotif}); and the signal is multivariate (two leads). Existing methods, at least in their current form, cannot recover all the occurrences of both motif sets (Fig. \ref{fig:ecg_existing}).

Motivated by problems of this type, we here propose an entirely new approach to TSMD that has none of the mentioned limitations. The proposed method, LoCoMotif, uses dynamic programming to find motif sets of varying length, measuring motif similarity using a time warping mechanism similar to DTW, in multivariate time series.\footnote{Source code for the method has been made available at \texttt{\url{https://github.com/ML-KULeuven/locomotif}} for reproducability.} LoCoMotif returns useful results on ECG data of the type just described (Fig.~\ref{fig:ecg_gt_locomotif}). Moreover, we will show that in a physiotherapy use case, our method successfully discovers repeated exercise executions in wearable sensor data. We also quantitatively evaluate LoCoMotif and compare it with existing methods, which is not common practice in the TSMD literature. To this end, we propose a way to construct labeled benchmark data from time series classification datasets, and also an evaluation metric to assess the quality of motif sets. On benchmark data, LoCoMotif finds better motif sets than existing methods, with comparable computational efficiency.

The remainder of this paper is organized as follows: after providing an overview of the related work (Section~\ref{sec:related_work}), we introduce our proposed method, called LoCoMotif (Section~\ref{sec:proposed}).  Afterwards, we propose evaluation metrics for the TSMD task (Section~\ref{sec:metrics}), which we use to evaluate LoCoMotif on a physiotherapy use case (Section~\ref{sec:usecase}) and to fairly compare the performance of LoCoMotif to existing methods on benchmark datasets (Section~\ref{sec:experiments}).
 
\section{Related Work}
\label{sec:related_work}
This section discusses the state-of-the-art TSMD methods are capable of handling at least one of the four challenges described in Section~\ref{sec:introduction} (more than two occurrences, variable-length, time warping, multivariate).%
\footnote{%
We do not consider methods that reduce time series data to a symbolic representation
because, 
to prevent the loss of relevant information for motif discovery, the discretization step requires carefully selected hyperparameters~\citep{mstamp,mueen2015enumeration,torkamani2017survey}, which are difficult to reason about~\citep{luo2012}. 
}
Table \ref{tab:properties} provides an overview of the capabilities, time complexity, and hyperparameters of these methods.

\newcommand{\rot}[1]{\rotatebox{90}{#1}}
\begin{table}
\begin{adjustbox}{width=\textwidth}
    \huge
    \begin{tabular}{llcccll}
        Method                              & \rot{Motif Sets ($k \geq 2$)}   & \rot{Variable-length}              & \rot{Time warping} & \rot{Supp. for multivar.} & Time Complexity & Hyperparameters \\ \midrule
        SetFinder     \citep{bagnall_finding_2014} & \cmark              & \xmark & \xmark & \xmark  & $O(n^2l)$ & $l, r$ \\
        LatentMotifs  \citep{grabocka_latent_2017} & \cmark & \xmark & \xmark & \xmark & $O(nl)$   & $l, r$ \\
        mSTAMP        \citep{mstamp}               & \xmark & \xmark & \xmark & \cmark & $O(n^2)$ & $l$  \\
        VALMOD        \citep{linardi_matrix_2018}  & \cmark & \cmark & \xmark & \xmark & $O(n^2(l_{\max}-l_{\min}))$ & $l_{\min}, l_{\max}, r_f$ \\
        SWAMP         \citep{alaee_matrix_2020}	& \xmark & \xmark & \cmark & \xmark & $O(n^2l^2)$ & $l$\\  
        Motiflets     \citep{motiflets}            & \cmark & \xmark & \xmark & \xmark 	& $O\left((l_{\max}-l_{\min})(k_{\max}n^2 + nk^2_{\max}\right))$ & $l_{\min}, l_{\max}, k_{\max}$ \\
        LoCoMotif                       & \cmark & \cmark & \cmark & \cmark & $O \left( \frac{n^2}{l_{\min}}(l_{\max} - l_{\min}) \right)$ & $l_{\min},l_{\max}, \rho$; optionally $\nu$ \\ \bottomrule
    \end{tabular}
\end{adjustbox}
\caption{Overview of capabilities, time complexity and hyperparameters of the considered TSMD methods. The stated time complexity is to discover one motif set.}
\label{tab:properties}
\end{table}

Many of the existing approaches use z-normalized Euclidean distance (ZED) as a similarity measure,  which implies they can compare only subsequences of identical length.  Among these, SetFinder \citep{bagnall_finding_2014} and LatentMotifs \citep{grabocka_latent_2017} are perhaps the most straightforward approaches: given hyperparameters $l$ and $r$, they evaluate each subsequence of length $l$ by counting the number of subsequences that are within a certain ZED radius $r$. Arguing that the need to specify $l$ limits the usability of the method and $r$ is very hard to specify for the user, \citet{motiflets}  present Motiflets.  Motiflets uses an algorithm that, for a given $k$ and $l$, finds the set of $k$ subsequences with length $l$ with smallest maximum pairwise ZED (avoiding the need to choose an $r$), and runs this algorithm for multiple values of $k$ and $l$ with $2 \leq k \leq k_{\max}$ and $l_{\min} \leq l \leq l_{\max}$ (for hyperparameters $l_{\min}$, $l_{\max}$ and $k_{\max}$). While Motiflets tries multiple values for $l$, it selects the most promising one; hence, all returned motif sets contain motifs of the same length.

VALMOD \citep{linardi_matrix_2018} allows motif length to vary among (not within) motif sets. For every subsequence with length within a given length range, it finds the most similar subsequence under ZED.  It next extends the pairs with the smallest ZED into larger sets by adding, for a pair with ZED $d$, all subsequences within radius $r=r_{f} \cdot d$ of the subsequences in the pair ($r_{f}$ is a hyperparameter). By first finding pairs and only secondarily sets, VALMOD is biased towards motif sets that contain a pair with very small ZED, disregarding other quality criteria.

SWAMP \citep{alaee_matrix_2020} is the only existing TSMD method that employs time warping: it uses z-normalized DTW instead of ZED. To limit the computation time, it only discovers one motif pair (the one with smallest DTW distance) and uses a fixed length $l$ (which is somewhat inconsistent with the motivation for using time warping).

mSTAMP \citep{mstamp} is the most recent method that addresses TSMD for multivariate time series. The authors consider the setting where some dimensions of the given time series are irrelevant for the TSMD task. For this purpose, mSTAMP first selects the relevant dimensions, and then returns the pair of subsequences of length $l$ that minimizes ZED over these dimensions.

While some of the above methods address some of the limitations of other methods, none of them meet all four criteria. It is also not trivial to extend them, as their limitations are often inherent to how the methods work. As will become clear, the proposed LoCoMotif method works very differently, meets all four criteria, and still scales quadratically with respect to the length $n$ of the input time series (Table~\ref{tab:properties}). It has hyperparameters: $\rho \in [0, 1]$, which controls how similar we want the subsequences in a discovered motif set to be; $l_{\min}$ and $l_{\max}$, which bound the length of the motifs to be discovered; and an optional hyperparameter $\nu \in [0, 0.5]$, which controls the allowed overlap within and across motif sets. \\
 
\noindent \textbf{Relation to Audio Thumbnailing Procedure (ATP)} LoCoMotif is inspired by ATP \citep{6353546}, which aims to find a representative segment for a given music recording; that is, an audio `thumbnail'. Since the authors define an audio thumbnail as a segment with many approximate repetitions that cover large parts of the music recording, ATP essentially finds a motif set. However, ATP's time complexity of $O(n^4)$ precludes its use as a general motif discovery method.  LoCoMotif crucially differs from ATP in that it uses a novel dynamic programming algorithm with complexity $O(n^2)$ once and extracts all motif sets from the result, while ATP applies such an algorithm for every possible segment in the time series.

\section{Proposed Method}\label{sec:proposed}
Before describing LoCoMotif, we introduce some background terms and concepts.

\subsection{Notation, Terminology and Auxiliary Concepts}\label{sec:terminology}

A \emph{time series} $\mathbf{x}$ of length $n$ is defined as a sequence of $n$ samples from a feature space $\mathcal{F}$: 
\begin{equation*}
    \mathbf{x} = (x_1, x_2, \dots, x_n)  \quad \text{where} \quad x_i \in \mathcal{F} \ \text{for} \ i = 1, \dots, n
\end{equation*}
$\mathcal{F}=\mathbb{R}$ for \emph{univariate} time series and $\mathcal{F}=\mathbb{R}^{d}$ for \emph{multivariate} time series with $d$ dimensions.
A \emph{segment} $\alpha$ that starts at index $b$ and ends at index $e$ is defined as:
\begin{equation*}
    \alpha = [b:e] = [b, b+1, \dots, e] \quad  \text{where} \quad  1 \leq b \leq e \leq n.
\end{equation*}
The length of segment $\alpha$ is denoted as $|\alpha|$ and is equal to $e-b+1$. A segment $\alpha$ relates to a \emph{subsequence} $\mathbf{x}_{\alpha}$, which consist of the time samples of $\mathbf{x}$ at the indices of $\alpha$. We define a segment $\alpha$ to be \emph{$\nu$-coincident} to another segment $\beta$ if $|\alpha \cap \beta| > \nu \cdot |\beta|$ where $\nu \in [0, 0.5]$ is the \emph{overlap} parameter. \\

\begin{figure}
	\begin{subfigure}[b]{0.49\linewidth}
		\centering
		\includegraphics[scale=0.42]{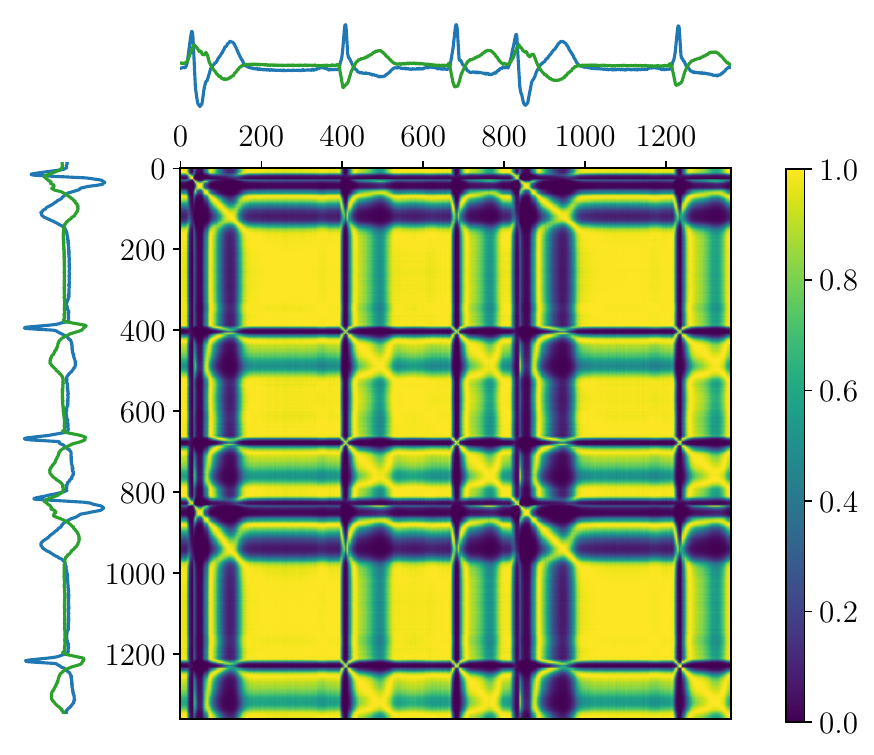}
		\caption{}
		\label{fig:ssm}
	\end{subfigure}
	\hfill
	\begin{subfigure}[b]{0.49\linewidth}
		\centering
		\includegraphics[scale=0.42]{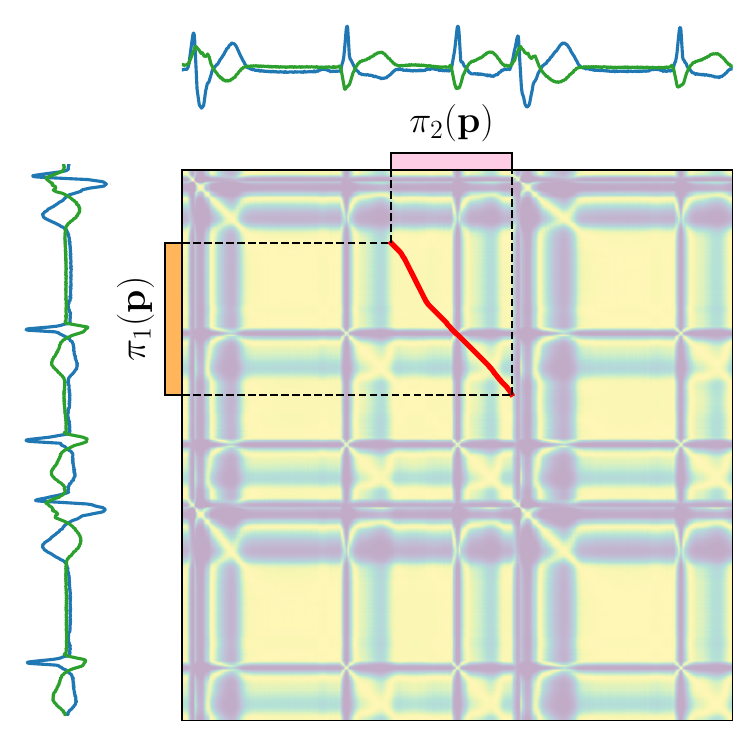}
		\caption{}
		\label{fig:ssm_proj}
	\end{subfigure}
	\caption{(a) The SSM of a part of the ECG example in Fig.~\ref{fig:ecg}. (b) A local warping path $\mathbf{p}$ represents a relation between its projections $\pi_1(\mathbf{p})$ and $\pi_2(\mathbf{p})$.
	}
\end{figure}

\noindent \textbf{Self-Similarity Matrix (SSM).} The \emph{SSM} contains the similarity between every pair of time samples in a given z-normalized time series $\mathbf{x}$ of length $n$. Using any similarity measure $s:\mathcal{F} \times \mathcal{F} \xrightarrow{} \mathbb{R}$, the SSM of $\mathbf{x}$ is defined as the square and symmetric matrix $\mathbf{S} = [S_{i, j}] = [s(x_i, x_j)]$ for every \emph{position} $(i, j) \in [1:n]^2$. We use the similarity measure $s(x, y)=\exp{(-\|x-y\|^{2})} \in [0, 1]$ (where $\|\cdot\|$ denotes the Euclidean norm) which is defined for both univariate and multivariate time series. 

The SSM $\mathbf{S}$ of $\mathbf{x}$ exposes its underlying structure and can be visualized to identify subsequences that are similar to each other. Fig.~\ref{fig:ssm} shows the SSM of a part of the ECG example (Fig.~\ref{fig:ecg_example}). A subsequence that approximately occurs elsewhere in the time series (in this case, a heartbeat) appears as a diagonal stripe of high similarity in $\mathbf{S}$. Such a stripe may exhibit curvature (i.e., have non-diagonal parts), indicating non-linear variations in time between the subsequence and its approximate occurrence. \\

\noindent \textbf{Local Warping Paths.} Such stripes of high-similarity can be captured using (possibly curved) local warping paths. Given an SSM $\mathbf{S} \in \mathbb{R}^{n \times n}$ of $\mathbf{x}$, a \emph{local warping path} of length $l$ is defined as a sequence $\mathbf{p} = (p_1, p_2, \dots, p_l)$ of positions $p_k = (i_k, j_k) \in [1:n]^2$ for $k \in [1:l]$ which is subject to the step size condition \citep{Mller2015}:
\begin{equation*}
	\label{eq:step_size}
	p_{k+1} - p_{k} \in \mathcal{A} \quad \text{for} \quad k \in [1:l-1]
\end{equation*}
where $\mathcal{A} = \{(1, 1), (1, 2), (2, 1)\}$ is \emph{the set of admissible steps}, constraining the slope of a path to lie between $\frac{1}{2}$ and 2.\footnote{A local warping path is a generalized version of the warping path defined in DTW as it is not subject to the boundary condition (Chapter~3 of~\citet{Mller2015}). While DTW finds a warping path that minimizes the distance (i.e., the dissimilarity), we find local warping paths that maximize similarity.} 
Each position $(i, j) \in \mathbf{p}$ maps the $i$th time sample of $\mathbf{x}$ to its $j$th time sample. Hence, a local warping path establishes a relation between two segments of $\mathbf{x}$, referred to as its \emph{projection} onto the vertical and horizontal axis, respectively defined as $\pi_1(\mathbf{p}) = [i_1:i_l]$ and $\pi_2(\mathbf{p}) = [j_1:j_l]$ (Fig.~\ref{fig:ssm_proj}). 

A local warping path relates the progression of time in both its projections. Along a path $\mathbf{p}$ in the forward direction (from $p_1$ to $p_l$), time increases monotonically in both projections. Diagonal steps on $\mathbf{p}$ represent an identical time progression in both projections, while non-diagonal steps indicate a temporal difference between them. Therefore, because of non-diagonal steps in $\mathcal{A}$, two subsequences that exhibit non-linear variations in time can successfully be related. Because of the slope constraint imposed by $\mathcal{A}$, the lengths of the projections of $\mathbf{p}$ differ at most by a factor of two.

\subsection{LoCoMotif}\label{sec:locomotif}

\begin{figure}
    \begin{subfigure}[b]{0.26\linewidth}
        \includegraphics[height=4cm]{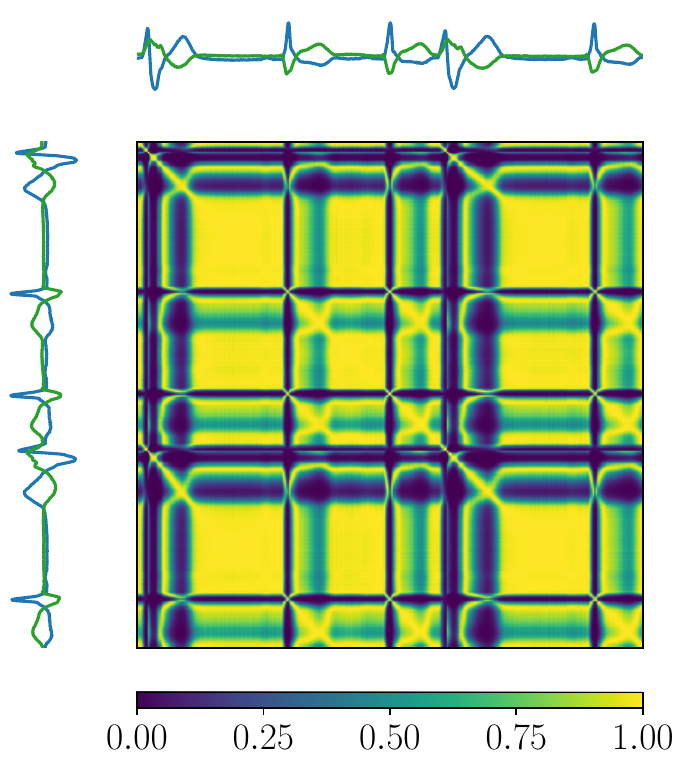}
        \caption{$\mathbf{S}$}\label{fig:overview_S}
    \end{subfigure}
    \hfill
    \begin{subfigure}[b]{0.22\linewidth}
        \includegraphics[height=4cm]{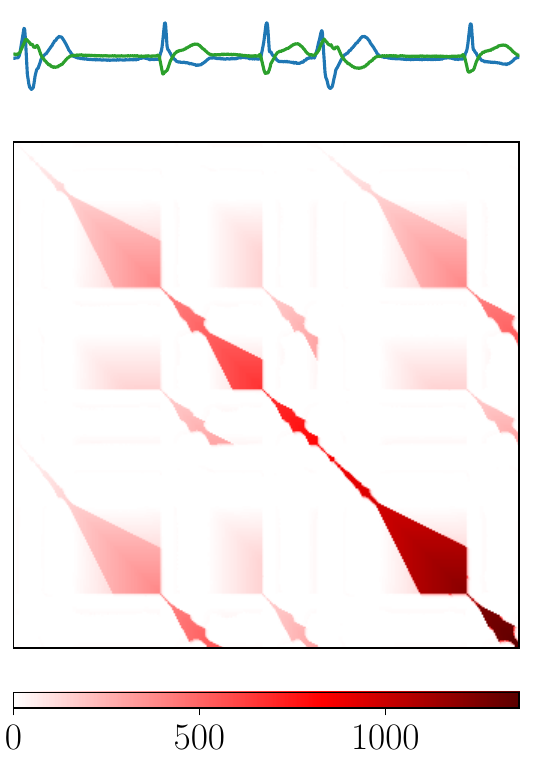}
        \caption{$\mathbf{D}$}\label{fig:overview_D}
    \end{subfigure}
    \begin{subfigure}[b]{0.22\linewidth}
        \includegraphics[height=4cm]{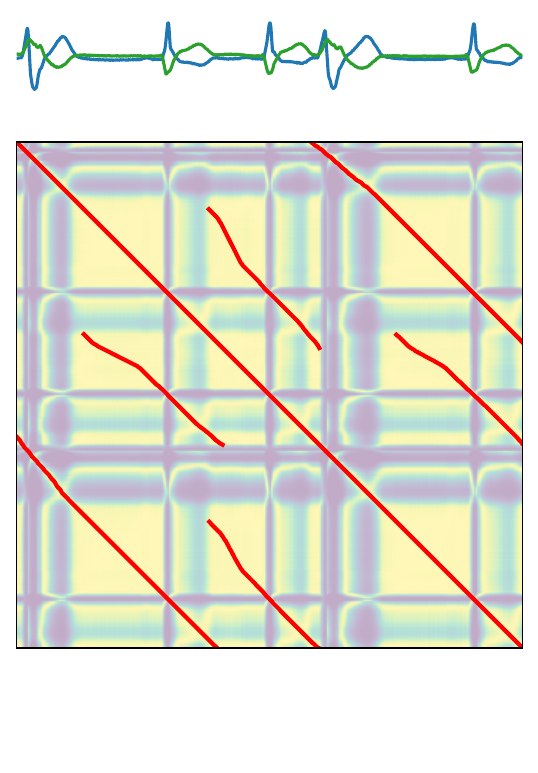}
        \caption{$\mathcal{P}$}\label{fig:overview_P}
    \end{subfigure}
    \begin{subfigure}[b]{0.25\linewidth}
        \includegraphics[height=4cm]{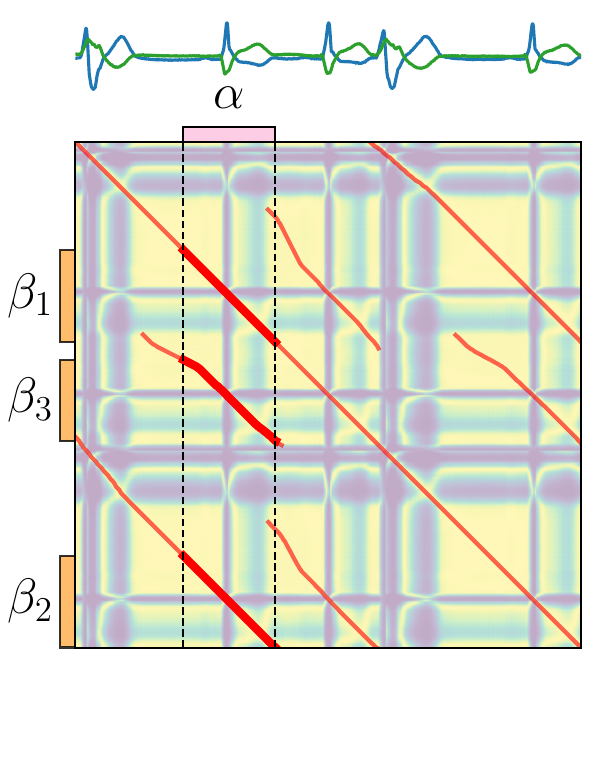}
        \caption{$\mathcal{M}_{\alpha} =\{\beta_1, \beta_2, \beta_3\}$}
        \label{fig:overview_M}
    \end{subfigure}
    \caption{Given a time series $\mathbf{x}$, LoCoMotif first calculates its SSM $\mathbf{S}$ (a), computes from this a ``cumulative similarity'' matrix $\mathbf{D}$ (b), and extracts a set of warping paths $\mathcal{P}$ from $\mathbf{D}$ (c). In a second step, LoCoMotif repeatedly uses $\mathcal{P}$ to associate with every candidate segment $\alpha$ a motif set $\mathcal{M_{\alpha}}$ (d).}
    \label{fig:overview}
\end{figure}

We first provide a high-level view of LoCoMotif, using Fig.~\ref{fig:overview} as a visual aid. The method consists of two steps, which are explained in detail in the next subsections.

In the first step, a novel algorithm called Local Concurrences (LoCo) calculates the SSM $\mathbf{S}$ (visualised in Fig.~\ref{fig:overview_S}), and from that, using dynamic programming, a ``cumulative similarity matrix'' $\mathbf{D}$ (Figure~\ref{fig:overview_D}) that contains in position $(i,j)$ the highest value that can be obtained by accumulating similarity values along any local warping path ending in position $(i,j)$ (using a specific aggregation function, which we detail later). LoCo extracts from $\mathbf{D}$ the paths with highest accumulated similarity (the red lines in Fig.~\ref{fig:overview_P}) and returns those in a set $\mathcal{P}$.


In the second step, LoCoMotif considers all segments $\alpha$ within a certain length range. For each of them, it finds all subpaths $\mathbf{q}$ of any path $\mathbf{p} \in \mathcal{P}$ for which $\pi_2(\mathbf{q})=\alpha$ (see Fig.~\ref{fig:overview_M}). Each such subpath $\mathbf{q}$ maps $\alpha$ to a segment $\beta = \pi_1(\mathbf{q})$. The set of all such $\beta$ segments, denoted $\mathcal{M}_{\alpha}$, is a candidate motif set.  LoCoMotif evaluates all candidates using some quality score, and adds the highest scoring one to its result set. It next repeats this step, ignoring segments that overlap too much with already discovered motifs.  It keeps repeating this until $\kappa$ motif sets have been discovered or no more can be found.

The following subsections describe the two steps in more detail.
\subsubsection{Local Concurrences (LoCo)}\label{sec:loco}
Local Concurrences (LoCo) calculates the SSM $\mathbf{S}$ of the input time series $\textbf{x}$ and finds a set of local warping paths $\mathcal{P}$ that covers the high-similarity regions of $\mathbf{S}$ (as shown in Fig.~\ref{fig:overview_P}). 
It does so by finding the paths along which the highest amount of similarity can be accumulated, and that do not contain prolonged \emph{gaps}, defined as consecutive low-similarity positions \citep{Mller2015}.

LoCo associates with every possible local warping path in $\mathbf{S}$ a value, called aggregated similarity, which is computed by traversing the path and accumulating the similarity in high-similarity positions, while strongly reducing the previously accumulated similarity when going through low-similarity positions. LoCo selects the local warping paths that maximize this value, since such paths have long stretches of high similarity and rarely contain gaps.

 The \emph{aggregated similarity} $\sigma(\mathbf{p})$ of a local warping path $\mathbf{p} = ((i_1, j_1), \dots, (i_l, j_l))$, is calculated as
\begin{equation}
	\sigma(\mathbf{p}) = (f_{i_l, j_l} \circ \dots \circ f_{i_1, j_1})(0)
\end{equation}
where the position-dependent function $f_{i, j}$ is defined as
\begin{equation}
	f_{i, j}(\sigma') = \begin{cases}
		\hphantom{\max(0, \delta_{m}}\sigma' \ + \ S_{i, j} & \text{if} \quad S_{i, j} \geq \tau \\
		\max(0, \delta_{m}\sigma' \ - \ \delta_{a})& \text{otherwise} \\
		\end{cases}
	\label{eq:f}.
\end{equation}
The function $f_{i,j}$ clearly implements the mechanisms described above: If the similarity $S_{i,j}$ at the current position $(i, j)$ is above a similarity threshold $\tau$ (i.e., of high similarity), $S_{i,j}$ is added to previously aggregated similarity $\sigma'$. Otherwise, $\sigma'$ is penalized in both an additive ($\delta_a \in \mathbb{R}^{+}$) and multiplicative ($\delta_m \in [0, 1]$) manner. When $\sigma'$ becomes 0, this means a sufficiently long gap (region of low similarity) has been encountered that it is better to start a new path from scratch at or after this point. 

\begin{figure}
\begin{subfigure}{0.49\linewidth}
    \centering
    \includegraphics[scale=0.45]{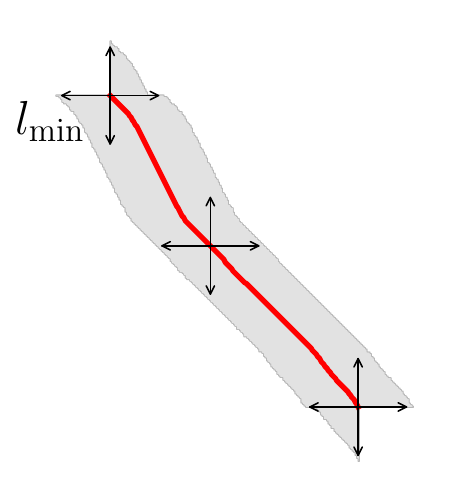}
    \caption{}
\end{subfigure}
\hfill
\begin{subfigure}{0.49\linewidth}
    \centering
    \includegraphics[scale=0.45]{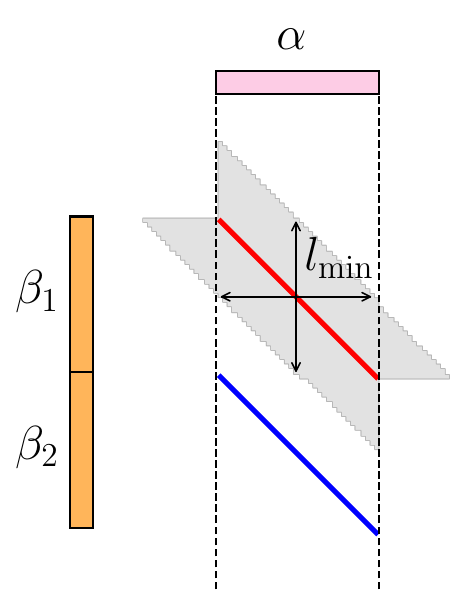}
    \caption{}\label{fig:vicinity_expl}
\end{subfigure}
\caption{(a) The vicinity of $\mathbf{p}$, with which subsequently found local warping paths cannot overlap, is defined as the union of cross-shaped regions of width $l_{\min}$ at every position of $\mathbf{p}$. (b) The extreme case where a segment $\alpha$ is related to two subsequent segments $\beta_1$ and $\beta_2$ (all of length $l_{\min}$).}
\label{fig:vicinity}
\end{figure}

To find the local warping paths that maximize aggregated similarity, LoCo uses dynamic programming (Algorithm \ref{algo:lc}). First, it calculates a matrix $\mathbf{D}$ that contains, for each position, the highest possible aggregated similarity that can be obtained by a local warping path $\mathbf{p}$ ending at that position: $D_{i,j} = \max\{\sigma(\mathbf{p}) \mid p_l = (i, j)\} \text{ for } (i, j) \in [1:n].$
The first value of $\mathbf{D}$ is initialized with
\begin{equation}
	D_{1, 1} =  f_{1,1}(0).
	\label{eq:LoCo_init}
\end{equation}
The highest aggregated similarity for any other position $(i,j)$ is calculated by extending the highest value among all positions that are reachable from $(i, j)$ using the admissible steps in $\mathcal{A}$ (Section \ref{sec:terminology}) in the backwards direction. For a position $(i,j)$, these positions are defined as
\begin{equation}
    \mathcal{I}_{i, j} =  \{(i - v, j - h) \mid (v, h) \in \mathcal{A}, i - v > 0, j - h > 0 \}
    \label{eq:admissable_positions}
\end{equation}
and the corresponding values as $\mathfrak{D}_{i, j} = \{D_{i', j'} \mid (i', j') \in \mathcal{I}_{i,j} \}$.
The remaining values in $\mathbf{D}$ can therefore be computed using:\footnote{We define $\max \emptyset = 0$.}
\begin{equation}
    D_{i, j} =  \ f_{i,j}(\max \mathfrak{D}_{i, j}).
    \label{eq:LoCo}
\end{equation}
The largest values in $D_{i,j}$ indicate the local warping paths that aggregate the largest amount of similarity. LoCo reconstructs these local warping paths one by one, imposing a certain space between them such that they are sufficiently distinct. More specifically, after reconstructing a path, LoCo masks the positions in its \emph{vicinity}, excluding them from subsequently found paths. The vicinity of a local warping path $\mathbf{p}$ is defined as the union of cross-shaped regions of width $l_{\min}$ centered at every position of $\mathbf{p}$ (as shown in Fig.~\ref{fig:vicinity}). The shape of the vicinity is motivated at the end of this section. 

\begin{algorithm}
\caption{\textsc{LoCo} (Local Concurrences)}\label{algo:lc}
\SetAlgoLined
\DontPrintSemicolon
\SetFuncSty{textsc}
\SetKwProg{myalg}{algorithm}{}{}
\SetKwFunction{LC}{LocalConcurrences}
\SetKwFunction{LWP}{Backtrack}
\SetKwInOut{KwIn}{Input}
\SetKwInOut{KwOut}{Output}
\KwIn{Time series $\mathbf{x}$, min.\ motif length $l_{\min}$, strictness parameter $\rho$}
\KwOut{Set of local warping paths $\mathcal{P}$}
Calculate the SSM $\mathbf{S}$ of $\mathbf{x}$.\;
Calculate $\mathbf{D}$ (Equation~\ref{eq:LoCo_init}, \ref{eq:admissable_positions} and \ref{eq:LoCo}) with $\tau$ set to $\rho$-quantile of $\mathbf{S}$.\;
$\mathcal{P} \gets \emptyset$\;
$\mathcal{X} \gets \emptyset$ \tcp{The set of all masked positions.}
\While(\tcp*[h]{LoCo stops when all indices are masked.}){$\mathcal{X} \neq [1:n]^{2}$}{
    \tcp{Find the unmasked position that has the highest value in $\mathbf{D}$:}
    $(i^{*}, j^{*}) \gets \arg \max_{(i, j)\in\mathcal{I} \setminus \mathcal{X}} D_{i, j}$\;
    \tcp{Reconstruct the path that ends at this position:}
    $\mathbf{p} \gets$ \LWP{$i^{*}$, $j^{*}$, $\mathbf{D}$, $\mathcal{X}$}\;
    \tcp{If the path is long enough, add the path to $\mathcal{P}$ and mask its vicinity; otherwise, only mask its positions:}
    \uIf{$|\pi_1(\mathbf{p})| \geq l_{\min}$ \Or $|\pi_2(\mathbf{p})| \geq l_{\min}$}{ 
        $\mathcal{X} \gets$ $\mathcal{X} \cup \{(i, j) \mid (i, j) \text{ in the vicinity of } \mathbf{p}\}$ \tcp{See Fig. \ref{fig:vicinity}.}
        $\mathcal{P} \gets \mathcal{P} \cup \{\mathbf{p} \}$\;
    }
    \Else{
        $\mathcal{X} \gets \mathcal{X} \cup \mathbf{p}$\;
    }
}
\KwRet $\mathcal{P}$\;
\vspace{1mm} \hrule \vspace{1mm}
\SetKwProg{myproc}{procedure}{}{}
\nonl \myproc{\LWP{$i$, $j$, $\mathbf{D}$, $\mathcal{X}$}}{
    $\mathbf{p} \gets ()$\;
    \tcp{LoCo stops backtracking when a zero or a masked position is encountered:}
    \While {$(i, j) \notin \mathcal{X}$ \And $D_{i, j} \neq 0$}{
        \tcp{Prepend the current position to the current path:}
        $\mathbf{p} \gets ((i, j), \ \mathbf{p})$\;
        \tcp{Obtain all positions that are admissible by $\mathcal{A}$:}
         $\mathcal{I} \gets \{(i-v,j-h) \mid (v, h) \in \mathcal{S}, i - v > 0, j - h > 0\}$\;
         \tcp{If there is no such position, a border of $\mathbf{D}$ is reached. In this case, stop backtracking:}
        \lIf{$\mathcal{I} = \emptyset$}{\Break}
        \tcp{Continue in the position that led to the value in the current position:}
        $(i, j) \gets \arg \max_{(i', j') \in \mathcal{I}} D_{i', j'} $\;
    }
    \KwRet $\mathbf{p}$\;
}
\end{algorithm}

LoCo continues until all positions have been masked (line 5). In each iteration, the best path is reconstructed using the \textsc{Backtrack} procedure, starting from the global maximum $(i^*, j^*)$ among all unmasked positions (line 6-7). The \textsc{Backtrack} procedure reconstructs the corresponding path by adding the current position to the path (line 16), and continuing the procedure recursively in the $(i', j') \in \mathcal{I}_{i,j}$ with the highest $D_{i', j'}$ (lines 17-19). This procedure stops when the current position contains a zero (indicating a prolonged gap) or when it is masked (i.e., it is on a previously obtained path) (line 15). After reconstructing a path, its vicinity is masked (line 9). We reject paths that are too short by discarding a path if it does not relate to any segment with a length greater than $l_{\min}$. In this case, we still mask the positions of the path, but not its vicinity (line 12), as this path is not included in the solutions.

The extent to which LoCo tolerates gaps can be controlled through the threshold $\tau$ and penalties $\delta_a$ and $\delta_m$. We set $\tau$ equal to the $\rho$-quantile of the SSM $\mathbf{S}$ and use $\delta_a = 2\tau$ and $\delta_m = 0.5$. Therefore, the user needs to select only the hyperparameter $\rho \in [0, 1]$, which has the default value of 0.8.

To motivate the shape of a path's vicinity, we consider the extreme case where two local warping paths relate a segment $\alpha$ (of length $l_{\min}$) to the two subsequent segments $\beta_1$ and $\beta_2$ (also of length $l_{\min}$) without time warping, as shown in Fig. \ref{fig:vicinity_expl}. In this case, the separation between the two local warping paths (blue and red) is $l_{\min}$. We allow this separation to get as small as $\frac{l_{\min}}{2}$ (both in the horizontal and vertical direction), so that $\beta_1$ and $\beta_2$ can overlap with each other to some extent, and the local warping paths can be curved to enable time warping.

\subsubsection{Discovering motif sets}\label{sec:discovering}
We now describe how LoCoMotif discovers motif sets based on the result of LoCo. To begin, we explain how LoCoMotif associates with every candidate $\alpha$ ($l_{\min} \leq |\alpha| \leq l_{\max}$) a motif set $\mathcal{M}_{\alpha}$ based on the set of local warping paths $\mathcal{P}$, and how LoCoMotif assesses the quality of such an $\mathcal{M}_{\alpha}$. \\

\begin{figure}
    \centering
    \includegraphics[scale=0.48]{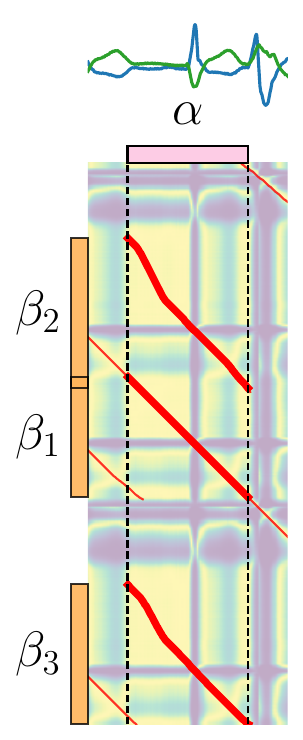}
    \caption{The set of subpaths $\mathcal{P}_{\alpha}$ (bold, red) and the candidate motif set $\mathcal{M}_{\alpha} =\{ \beta_1, \beta_2, \beta_3 \}$ of the candidate segment $\alpha$.}
    \label{fig:matches}
\end{figure}

\noindent \textbf{Candidate motif set.} As shown in Fig.~\ref{fig:matches}, the \emph{candidate motif set} $\mathcal{M}_{\alpha}$ of $\alpha$ is obtained by considering every subpath $\mathbf{q}$ of a path $\mathbf{p} \in \mathcal{P}$ that has $\alpha$ as its projection to the horizontal axis:
\begin{equation}
    \mathcal{P}_{\alpha} = \{ \mathbf{q} \mid \pi_2(\mathbf{q}) = \alpha \land \exists \mathbf{p} \in \mathcal{P}: \mathbf{q} \subseteq \mathbf{p} \in \mathcal{P} \}
    \label{eq:P}
\end{equation}
and then obtaining the segments related to $\alpha$ by $\mathcal{P}_\alpha$:
\begin{equation}
    \mathcal{M}_{\alpha} = \{ \pi_1(\mathbf{q}) \mid \mathbf{q} \in \mathcal{P}_{\alpha} \}.
    \label{eq:M}
\end{equation}
Note that the segments in $\mathcal{M}_{\alpha}$ can overlap (Fig.~\ref{fig:matches}); $\alpha \in \mathcal{M}_{\alpha}$, since the diagonal of $\mathbf{S}$ is guaranteed to be found as a local warping path (Fig.~\ref{fig:overview_S}); and because of time warping, the lengths of $\alpha$ and a segment in $\mathcal{M}_{\alpha}$ can differ, maximally by a factor of 2 because of the slope constraint imposed by $\mathcal{A}$ (Section \ref{sec:terminology}). \\

\noindent \textbf{Quality of a motif set.} To assess the quality of a candidate motif set, LoCoMotif uses the fitness measure of \citet{6353546}, consisting of two components: \emph{score} $\varsigma(\alpha)$, which measures how similar $\alpha$ is to the segments in $\mathcal{M}_{\alpha}$, and \emph{coverage} $\gamma(\alpha)$, which quantifies how much of the time series is covered by $\mathcal{M}_{\alpha}$:
\begin{equation*}
\varsigma(\alpha) = \sum_{\mathbf{q} \in \mathcal{P}_\alpha}{\sum_{(i, j) \in \mathbf{q}} S_{i, j}} \quad \text{and} \quad
      \gamma(\alpha) = \sum_{\beta \in \mathcal{M}_{\alpha}} |\beta| - \sum_{\beta_1, \beta_2 \in \mathcal{M}_\alpha, \beta_1 \neq \beta_2} |\beta_1 \cap \beta_2|.
\end{equation*}
To equalize the effects of the two components, both are normalized (after subtracting the contribution of the self-match $\alpha$): 
\begin{equation*}
\overline{\varsigma}(\alpha) = \frac{\varsigma(\alpha) - |\alpha|}{\sum_{\mathbf{q} \in \mathcal{P}_\alpha}{|\mathbf{q}|}}  \quad \text{and} \quad \overline{\gamma}(\alpha) = \frac{\gamma(\alpha) - |\alpha|}{n} 
\end{equation*}
such that $\overline{\varsigma}(\alpha), \overline{\gamma}(\alpha) \in [0, 1-\frac{|\alpha|}{n}]$.
The \emph{fitness} $\varphi(\alpha)$ of $\alpha$ is defined as the harmonic mean of the normalized coverage and score:
\begin{equation}
    \varphi(\alpha) = 2 \cdot \frac{\overline{\varsigma}(\alpha) \cdot \overline{\gamma}(\alpha)}{\overline{\varsigma}(\alpha) + \overline{\gamma}(\alpha)}.
    \label{eq:fitness}
\end{equation}
and is used to quantify the quality of the candidate motif set $\mathcal{M}_\alpha$. \\

We can now describe the full LoCoMotif method (Algorithm \ref{algo:locomotif}). First, it applies LoCo (Algorithm \ref{algo:lc}) to find the set of local warping paths $\mathcal{P}$ (line 1). Second, it iteratively uses the \textsc{BestMotifSet} procedure to find the candidate motif set with the highest fitness (line 4), until $\kappa$ motif sets are discovered (if $\kappa$ is specified) (line 3) or there is no good candidate motif set left (line 5).

\begin{algorithm}
\caption{LoCoMotif}\label{algo:locomotif}
\SetAlgoLined
\DontPrintSemicolon
\SetFuncSty{textsc}
\SetKwInOut{KwIn}{Input}
\SetKwInOut{KwOut}{Output}
\SetKwFunction{loco}{LoCo}
\SetKwFunction{bestmotifset}{BestMotifSet}
\SetKwProg{myproc}{procedure}{}{}
\KwIn{Time series $\mathbf{x}$, min.\ and max.\ length $l_{\min}$ and $l_{\max}$, strictness $\rho$, number of motif sets $\kappa$ (opt.), overlap $\nu$ (opt., 0.5 by default).}
\KwOut{Motif sets $\mathcal{M}$}
$\mathcal{P} \gets$ \loco{$\mathbf{x}, l_{\min}, \rho$}\; 
$\mathcal{M} \gets \emptyset$\;
\While{$|\mathcal{M}| < \kappa$}{
    $\mathcal{M}_{\alpha^{*}} \gets $ \bestmotifset{$\mathcal{P}$, $l_{\min}$, $l_{\max}$, $\mathcal{M}$}\;
    \lIf{$\mathcal{M}_{\alpha^{*}} = \emptyset$}{\Break}
    $\mathcal{M} \gets \mathcal{M} \cup \{ \mathcal{M}_{\alpha^{*}} \}$\;
}
\Return{$\mathcal{M}$}
\vspace{1mm} \hrule \vspace{1mm}
\nonl \myproc{\bestmotifset{$\mathcal{P}$, $l_{\min}$, $l_{\max}$, $\mathcal{M}$}}{
    $(\mathcal{M}_{\alpha^{*}}, \varphi_{\alpha^{*}}) \ \gets (\emptyset, 0)$\;
    $\mathcal{P}^b \gets \emptyset$ \tcp{$\mathcal{P}^b$ contains every $\mathbf{p}$ for which $b \in \pi_2(\mathbf{p}$)}
    \For {b \In $[1:n-l_{\min}]$}{
        $\mathcal{P}^b \gets \mathcal{P}^b \cup \{ \mathbf{p} \in \mathcal{P} \mid \text{start index of } \pi_2(\mathbf{p}) = b \}$\;
        $\mathcal{P}^b \gets \mathcal{P}^b \setminus \{ \mathbf{p} \in \mathcal{P}^b \mid \text{end index of } \pi_2(\mathbf{p}) = b-1 \}$\;
        $\mathcal{P}^e \gets \mathcal{P}^b$ \tcp{$\mathcal{P}^e$ contains every path for which $b, e \in \pi_2(\mathbf{p})$}
	\For {e \In $[b + l_{\min}-1:\min(b + l_{\max}-1, n)]$}{
            $\mathcal{P}^e \gets \mathcal{P}^e \setminus \{ \mathbf{p} \in \mathcal{P}^{e} \mid \text{end index of } \pi_2(\mathbf{p}) = e-1 \}$\;
		$\alpha \gets [b:e]$\;
		\If{\upshape $\alpha$ is $\nu$-coincident to any segment in $\mathcal{M}$}{\Break}
		Obtain $\mathcal{P}_{\alpha}$ and $\mathcal{M}_{\alpha}$ based on the paths in $\mathcal{P}^e$ (Equation \ref{eq:P} and \ref{eq:M})\;
		Discard the segments in $\mathcal{M}_{\alpha}$ that are $\nu$-coincident to any segment in $\mathcal{M}$ (also discard the corresponding paths in $\mathcal{P}_{\alpha}$).\;
		\If{\upshape $\exists \beta_1, \beta_2 \in \mathcal{M}_{\alpha}$: $\beta_1$ \text{is $\nu$-coincident to} $\beta_2$}{\Continue}
		$\varphi_{\alpha} \gets $ fitness of $\mathcal{M}_{\alpha}$ (Equation \ref{eq:fitness}).\;
            \If{$\varphi_{\alpha} > \varphi_{\alpha^{*}}$}{$(\mathcal{M}_{\alpha}^{*}, \varphi_{\alpha}^{*}) \gets (\mathcal{M}_{\alpha}, \varphi_{\alpha})$\;}		
        }
    }
    \Return $\mathcal{M}_{\alpha^{*}}$
}
\end{algorithm}

The \textsc{BestMotifSet} procedure calculates the fitness for every candidate motif set, and returns the one that maximizes fitness. For efficiency, the fitness values are calculated in an incremental manner: The procedure loops over every possible start index $b$ of a segment $\alpha$ (line 10), increments its end index $e$ from $b+l_{\min}-1$ to $b+l_{\max}-1$ (line 14), while calculating the fitness of candidate $\alpha=[b:e]$. During this process, two data structures $\mathcal{P}^{b}$ and $\mathcal{P}^{e}$ are maintained, which are incrementally updated together with $b$ and $e$, respectively. $\mathcal{P}^{b}$ includes every path $\mathbf{p} \in \mathcal{P}$ for which $b \in \pi_2(\mathbf{p})$. Every time $b$ is incremented, every path $\mathbf{p} \in \mathcal{P}$ whose projection $\pi_2(\mathbf{p})$ starts with $b$ is added to $\mathcal{P}^{b}$ (line 11), and every $\mathbf{p}$ whose projection $\pi_2(\mathbf{p})$ ends with $b-1$  is removed from it (line 12). $\mathcal{P}^e$ is the subset of $\mathcal{P}^{b}$ that contains every path $\mathbf{p}$ such that $b, e \in \pi_2(\mathbf{p})$. For a new $b$, it is initialized to $\mathcal{P}^{b}$ (line 13), and updated together with $e$ by removing every path $\mathbf{p}$ whose $\pi_2(\mathbf{p})$ ends with $e-1$ (line 15). The motif set $\mathcal{M}_{\alpha}$ is obtained solely based on the paths in $\mathcal{P}^e$ (line 19), since $\mathcal{P}^e$ always consists of the paths that are relevant for the current $\alpha$.

Each call of the \textsc{BestMotifSet} procedure takes into account the previously discovered motif sets $\mathcal{M}$, such that LoCoMotif does not discover the same segment multiple times. To this end, we implement two measures: any candidate segment $\alpha$ that is $\nu$-coincident to a segment in $\mathcal{M}$ is not considered (lines 17-18) and any segment in candidate motif set $\mathcal{M}_{\alpha}$ that is $\nu$-coincident to a segment in $\mathcal{M}$ is discarded (line 20). The \textsc{BestMotifSet} procedure also ignores every $\mathcal{M_{\alpha}}$ that contains a pair of $\nu$-coincident segments (lines 21-22). This ensures that, when applied on a quasi-periodic time series, LoCoMotif discovers motifs that each contain one cycle, instead of multiple cycles per motif. The flexibility of these measures can be controlled through overlap parameter $\nu$.

\subsubsection{Optimizations and Time Complexity}
We now discuss how we optimize Algorithm \ref{algo:locomotif} such that the time complexity of one iteration is equal to $O\left(\frac{n^2}{l_{\min}}(l_{\max} - l_{\min})\right)$. Since LoCo has a time complexity of $O(n^2)$, it suffices to show that the \textsc{BestMotifSet} procedure has a time complexity of $O\left(\frac{n^2}{l_{\min}}(l_{\max} - l_{\min})\right)$. 

The inner loop of the \textsc{BestMotifSet} procedure (line 15-25) is executed $O(n(l_{\max} - l_{\min}))$ times; hence, we need to show that it has a time complexity of $O(\frac{n}{l_{\min}})$. To this end, we first argue that $\mathcal{P}^e$ contains $O(\frac{n}{l_{\min}})$ paths. Then, we show that the inner loop has a time complexity of $O(|\mathcal{P}^e|)$. 

For any $b$, it holds that $|\mathcal{P}^b| < \frac{2n}{l_{\min}}$, because LoCo imposes a minimum separation of $\frac{l_{\min}}{2}$ between any two found local warping paths (Section \ref{sec:loco}). Since $\mathcal{P}^e \subseteq \mathcal{P}^b$, $|\mathcal{P}^e| < \frac{2n}{l_{\min}}$ as well (in practice, $|\mathcal{P}^b|,|\mathcal{P}^e| \ll \frac{2n}{l_{\min}}$). 

Lastly, we need to show that every line of the inner loop has a time complexity of $O(|\mathcal{P}^{e}|)$. We consider only line 19 and lines 21-23, since line 15-16 are trivially $O(|\mathcal{P}^{e}|)$ and lines 17, 18 and 20 are not used in the first iteration because $\mathcal{M}=\emptyset$:
\begin{itemize}
    \item Line 19: To efficiently obtain $\mathcal{P}_{\alpha}$ and $\mathcal{M}_{\alpha}$ for any candidate $\alpha = [b:e]$, we associate with every path $\mathbf{p} \in \mathcal{P}$ a lookup table that contains where a certain $j$ occurs for the first time as a column index in $\mathbf{p}$. In mathematical terms, we associate with $\mathbf{p}$ a function $k_{\mathbf{p}}$, defined as $k_{\mathbf{p}}(j) = \min \{ k \mid (i_k, j_k) \in \mathbf{p}, j \geq j_k \}$ where we use $\geq$ instead of $=$ because $j$ may not be present in $\mathbf{p}$ due to the admissible steps in $\mathcal{A}$ that skip a column index. Using these lookup tables, both $\mathcal{P}_{\alpha}$ = $\{ \left(p_{k_\mathbf{p}(b)},p_{k_\mathbf{p}(b) + 1}, \dots, p_{k_\mathbf{p}(e)}\right) \mid \mathbf{p} \in \mathcal{P}^e \}$ and $\mathcal{M}_{\alpha}$ can be calculated in $O(|\mathcal{P}^e|)$ time.
    \item Lines 21-22: To check $\nu$-coincidence in $\mathcal{M}_{\alpha}$, the segments in $\mathcal{M}_{\alpha}$ are sorted by their start index, such that overlaps can be calculated by considering every two consecutive segments. To this end, the paths in $\mathcal{P}^b$ (and therefore also in $\mathcal{P}^e$) are sorted by $i_k$, where $(i_k, j_k)=p_{k_{\mathbf{p}}(b)}$.
    \item Line 23: To efficiently calculate the fitness value of $\alpha = [b:e]$, we optimize the calculation of score $\varsigma(\alpha)$ (coverage $\gamma(\alpha)$ is simply obtained using the lengths of the segments in $\mathcal{M}_{\alpha}$ and the overlaps between them). To this end, we use another lookup table for every path $\mathbf{p} \in \mathcal{P}$ which contains, for every $j \in \pi_2(\mathbf{p})$, the cumulative similarity from $p_1$ to $p_{k_{\mathbf{p}}(j)}$: $\varsigma_{\mathbf{p}}(j) = \sum_{1 \leq k \leq k_{\mathbf{p}}(j)} S_{p_k}$. Using these lookup tables, $\varsigma(\alpha) = \sum_{\mathbf{p} \in \mathcal{P}^e} \varsigma_{\mathbf{p}}(e) - \varsigma_{\mathbf{p}}(b)$ is also computed in $O\left(|\mathcal{P}^e|\right)$ time. 
\end{itemize}

\subsubsection{Variants of LoCoMotif}\label{sec:variants}
Lastly, we present two versions of LoCoMotif which may be useful in practical applications. \\

\noindent \textbf{LoCoMotif without time warping.} The time warping feature of LoCoMotif can be optionally disabled by using the set of admissible steps $\mathcal{A} = \{(1, 1)\}$. In this case, LoCo only finds strictly diagonal local warping paths and all motifs in a motif set have the same length. In our experiments (Section \ref{sec:usecase} and \ref{sec:experiments}), we use a binary hyperparameter to enable or disable time warping.  \\

\noindent \textbf{Guiding LoCoMotif.} In practice, users of a TSMD method typically have domain-dependent preferences towards certain motifs or constraints on motifs. To address this, an existing study~\citep{guiding} proposes the use of a user-specified time series (of length $n$) that quantifies the user's preference for a motif starting at a certain index.
LoCoMotif supports such a binary time series that allows the user to constrain the possible start and end points of a candidate segment. This feature is used in the physiotherapy use case in Section~\ref{sec:usecase}.

\section{Evaluation of TSMD methods}\label{sec:metrics}
In the literature, TSMD methods are typically evaluated in terms of computation time \citep{linardi_matrix_2018, mstamp, grabocka_latent_2017} or motif quality. For the latter, existing methods have mostly been evaluated qualitatively \citep{guiding, mstamp} or with metrics that do not cover all aspects of TSMD. For example, some evaluation metrics are only defined for a single motif set, or do not penalize false discoveries (i.e., undesirable segments in a discovered motif set) \citep{unitaod, motiflets}. To avoid such issues, we generalize the F1-score metric from classification to the TSMD task. 

Evaluating the quality of the discovered motif sets $\mathcal{M} = \{ \mathcal{M}^{(1)}, \dots, \mathcal{M}^{(\kappa)}\}$ of a TSMD method requires knowledge of the actual segments of the patterns present in the time series. These segments, which we assume to be non-overlapping,  are referred to as the \emph{ground truth} (GT) motif sets $\mathcal{G} = \{\mathcal{G}^{(1)}, \dots, \mathcal{G}^{(\kappa')}\}$. Our evaluation metric measures how well $\mathcal{M}$ corresponds to $\mathcal{G}$.

\begin{figure}
\centering 
\begin{subfigure}[b]{0.48\linewidth}
\centering
  \includegraphics[width=\linewidth]{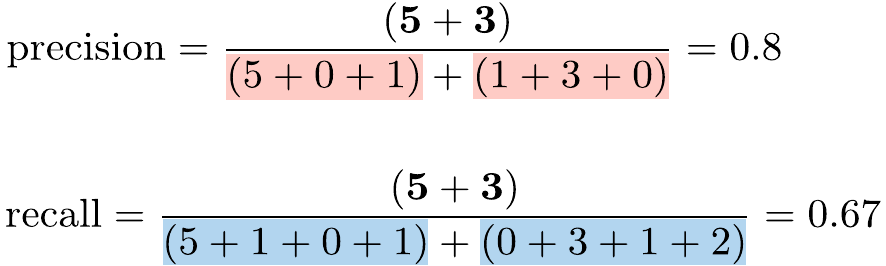}
  \caption{}
\label{fig:metrics_confusion_matrix}
\end{subfigure}
\begin{subfigure}[b]{0.48\linewidth}
\centering
  \includegraphics[width=\linewidth]{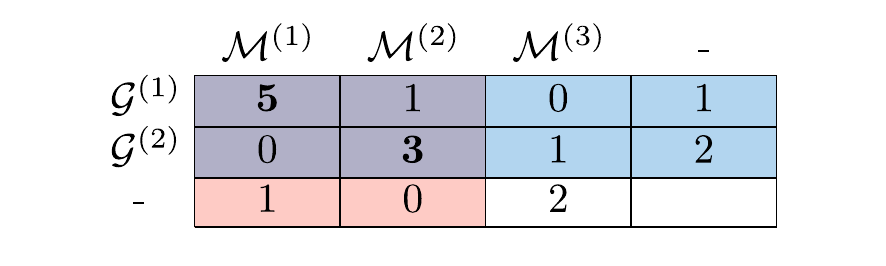}
  \caption{}
\label{fig:metrics_prec_rec}
\end{subfigure}
\caption{(a) An example matching matrix $\mathbf{M^{*}}$  (where the number of GT motif sets $\kappa'=2$ and the number of discovered motif sets $\kappa=3$), and (b) the computation of the corresponding precision and recall.}
\label{fig:metrics}
\end{figure}

First, we match every GT segment with at most one discovered segment and vice versa. More specifically, we match a GT segment $\beta$ with a discovered segment $\alpha$ when $\frac{|\alpha \cap \beta|}{|\alpha \cup \beta|} > 0.5$. If there are multiple such $\alpha$, we select the one with the highest ratio (and if there are multiple with highest ratio, we arbitrarily select one of them).
Due to the threshold of 0.5, at most one $\beta$ can be associated with the same $\alpha$.\footnote{Proof: $|\alpha \cap \beta| > 0.5 |\alpha \cup \beta| \geq 0.5 |\alpha|$, so $\beta$ must cover more than half of $\alpha$; if two GT segments $\beta$ and $\beta'$ fulfilled this condition, they would have to overlap, which contradicts the disjointness of GT segments.} The solution for a given set of discovered and GT segments is therefore unique up to equally-good matches (the arbitrary selection mentioned above).

Second, we assign the motif sets in $\mathcal{M}$ to the motif sets in $\mathcal{G}$. To this end, we construct a so-called \emph{matching matrix} $\mathbf{M}^{*} \in \mathbb{N}^{(\kappa'+1) \times (\kappa+1)}$, which is a kind of generalized confusion matrix. Consider $\mathbf{M} \in \mathbb{N}^{\kappa' \times \kappa}$, with $M_{i, j}$ the number of matched segments between $\mathcal{G}^{(i)}$ and $\mathcal{M}^{(j)}$. $\mathbf{M}^*$ is defined by first permuting the rows and columns of $\mathbf{M}$ such that the sum of its diagonal is maximized \citep{assignment}, then adding an additional column that indicates for each GT motif set how many of its segments were not matched with a discovered segment, and similarly adding a row for discovered segments not matched with a GT segment. See Fig.~\ref{fig:metrics_confusion_matrix} for an example.

To define F1-score, we use the micro-average of precision and recall for multi-class classification~\citep{grandini2020metrics}:
\begin{equation*}
\text{precision} =\frac{\sum_{i=1}^{\kappa_{\min}}{M_{i, i}^{*}}}{\sum_{j=1}^{\kappa_{\min}}{\left( \sum_{i=1}^{\kappa'+1}{M_{i, j}^{*}}\right)}} \quad \text{ and } \quad 
\text{recall} = \frac{\sum_{i=1}^{\kappa_{\min}}{M_{i, i}^{*}}}{\sum_{i=1}^{\kappa'}{\left( \sum_{j=1}^{\kappa+1}{M_{i, j}^{*}}\right)}}
\end{equation*}
where $\kappa_{\min} = \min(\kappa, \kappa')$ (see Fig.~\ref{fig:metrics_prec_rec} for an example). Our F1-score is defined as the harmonic mean of this precision and recall, and evaluates the quality of $\mathcal{M}$ in all aspects, as precision penalizes unmatched segments in discovered motif sets (that is, false discoveries) and recall penalizes undiscovered segments in GT motif sets. 

\section{Use Case: Physiotherapy}\label{sec:usecase}
In this section, we use LoCoMotif to solve a TSMD task that emerges in the field of physiotherapy.  \\

\noindent \textbf{Context.} In physiotherapy, there is interest in systems that automatically provide feedback to patients by using wearable motion sensors. Such systems identify every performed exercise execution and assess whether it is correctly executed or not. This task is performed typically in a supervised manner, relying on training data that consist of both correctly and incorrectly performed exercise executions, often acquired from every patient separately \citep{yurtman_automated_2014}. However, because exercises are repeatedly executed, they can also be discovered and grouped based on how they were performed in an unsupervised manner, which is a TSMD task. This unsupervised approach has the advantage of not requiring any training data. Although an expert is still needed to evaluate the correctness of the discovered exercise executions, this can be done much more efficiently by evaluating only a representative segment from each discovered motif set.

LoCoMotif stands out as the only applicable TSMD method to this use case because of its unique capabilities: it can be applied to the multivariate time series measured by the wearable motion sensors (whereas other methods cannot), and in the context of physiotherapy, LoCoMotif tolerates (to a limited extent) temporal variations in executions of the same exercise. \\
\begin{figure}
    \centering
    \includegraphics[width=\textwidth]{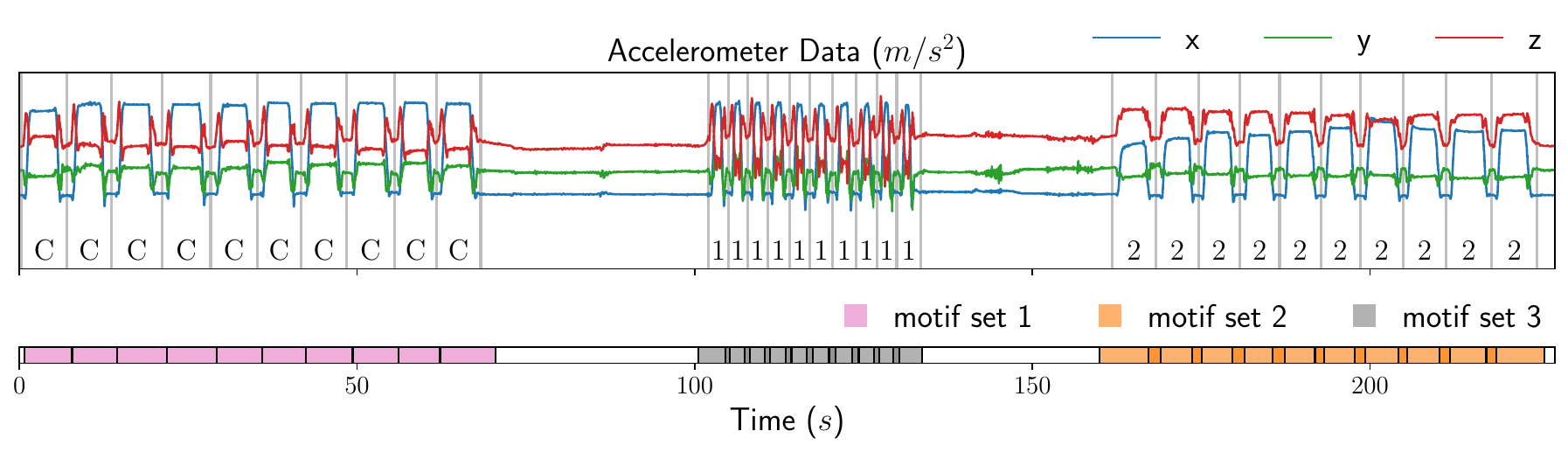}
    \caption{Each recording (here, subject 3 performing exercise 6) from the physiotherapy exercises dataset consists of three sections of ten exercise executions, interleaved with idle segments. LoCoMotif perfectly discovers the exercise executions in this example recording (F1-score of 1.00).}
    \label{fig:physical_therapy_example}
\end{figure}

\noindent \textbf{Dataset.} 
We use a publicly available dataset~\citep{physicaltherapydataset} where body movements were captured by five Inertial Measurement Units (IMUs) at 25~Hz. We only use the test set which contains 40 recordings of physiotherapy sessions (five subjects, each performing eight exercises).
Each recording contains 30 exercise executions of three different types (Fig.~\ref{fig:physical_therapy_example}): ten performed in a correct manner (C), ten performed too quickly (error type 1) and ten performed with a low range of motion (error type 2). For each exercise, we use the tri-axial accelerometer in the IMU that moves the most (IMU4 for exercise 2 and IMU2 for the other exercises). \\

\noindent \textbf{Goal.} The goal is to extract all exercise executions and separate them into three motif sets based on their type (C, error type 1, error type 2). \\

\noindent \textbf{Experimental Setup.} 
We apply LoCoMotif both with and without time warping (Section \ref{sec:variants}). We set $l_{\min} = 51$ and $l_{\max} = 262$ according to the length of the shortest and longest exercise execution in the whole dataset, extract $\kappa=6$ motif sets, and set the overlap parameter to $\nu=0.25$.\footnote{We allow LoCoMotif to discover more motif sets than there actually are because the expert can easily discard the undesired ones.}  To study the effect of hyperparameter $\rho$, we run the experiment for $\rho \in \{0.3, 0.4, \dots, 0.9\}$.

For this use case, we guide LoCoMotif (Section \ref{sec:variants}) and constrain the candidate segments to start and end at time indices where the subject is in resting position. To this end, we automatically identify the long, idle segments in the recording (e.g., the two intervals with almost constant acceleration in Fig.~\ref{fig:physical_therapy_example}) by using a sliding window of size $l_{\max}$ and taking the union of the windows where the variance in every dimension is below an absolute threshold. Then, we only consider candidate segments whose start and end points are among the $33\%$ of the time samples (outside the idle segments) that are closest to the mean value of the identified idle segments. \\
\begin{figure}
    \centering
    \includegraphics[scale=0.33]{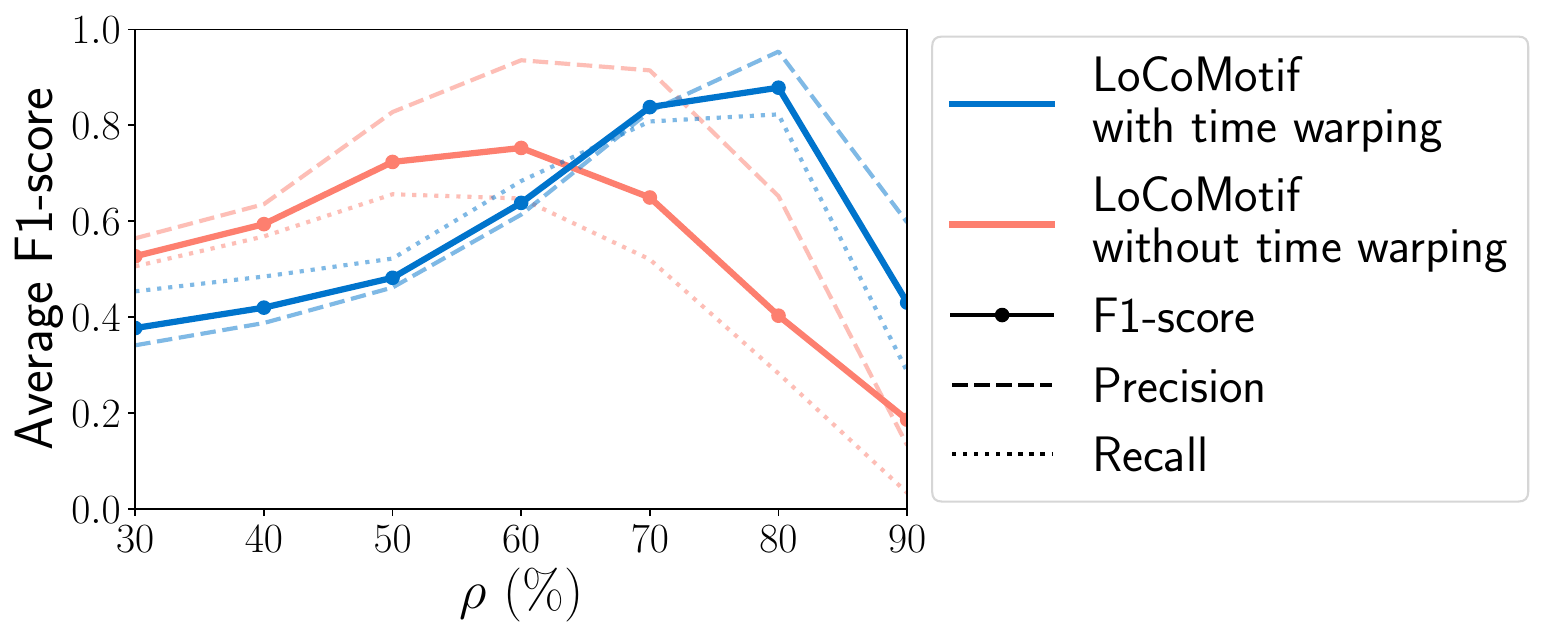}
    \caption{For the physiotherapy use case, LoCoMotif achieves a higher F1-score when it uses time warping, for higher values of $\rho$.}
    \label{fig:results_physicaltherapy}
\end{figure}

\noindent \textbf{Results.}  LoCoMotif achieves a high average F1-score of 0.88 using time warping (when $\rho=0.8$), which is higher than the one obtained without time warping (0.75 when $\rho = 0.6$). On average, LoCoMotif (with time warping, $\rho=0.8$) requires 10 seconds to process a physiotherapy recording that has an average duration of 220 seconds (5500 time samples), when ran on a computer with an Intel(R) Xeon(R) E3-1225 v3 CPU running at 3.20 GHz and 32 GB RAM.

The separation of exercise types into different motif sets depends on hyperparameter $\rho$: Selecting a value for $\rho$ that is too low ($0.3 \leq \rho \leq 0.5$) causes motif sets in which execution types are mixed, while a higher value ($0.6 \leq \rho \leq 0.8$) makes LoCoMotif more strict and causes exercise types to be
separated. Using a value that is too high ($\rho = 0.9$), however, prevents any exercise type from being discovered. Fig.~\ref{fig:physical_therapy_example} shows a time series in which the different types of executions are perfectly found as motif sets. \\

\noindent \textbf{Conclusion.} We conclude that LoCoMotif successfully completes the given task in a reasonable time. Using time warping allows LoCoMotif to be more strict (higher $\rho$), while still finding the right motifs.
\section{Experimental Evaluation}\label{sec:experiments}
In this section, we systematically and quantitatively compare the performance of LoCoMotif against existing TSMD methods using the F1-score metric of Section~\ref{sec:metrics}. We consider both a general setting where motifs in a motif set have different lengths, and a specific setting in which every motif of every motif set has the same length. For both settings, we construct benchmark datasets, using a procedure that generates time series intended for TSMD by concatenating instances of a dataset intended for classification (described in Section~\ref{sec:generation}).

\subsection{Benchmark Data Generation}\label{sec:generation} 
A common approach in the literature to create a TSMD benchmark dataset is to concatenate instances from a time series classification dataset. In this case, the time series instances that belong to the same class are assumed to be similar and are expected to be discovered as a motif set. In existing studies, time series instances are interleaved with random-walk segments \citep{unitaod}, which is not ideal since it is trivial to distinguish highly structured instances from randomly generated segments. To address this, we solely concatenate instances from the classification dataset, which makes our benchmarks more representative of realistic scenarios because no synthetic data is used.


We first describe how a time series that we generate looks. Consider a classification dataset with five classes: A, B, C, D and E. One could randomly concatenate instances from these classes; however, in this case a time series like \underline{AB}C\underline{AB}DE could occur, where AB may be found as a motif, instead of the expected motifs A and B. We solve this by imposing a certain structure on each time series we generate: between any two instances of any class that is repeated, we put a single instance from a class that does not repeat, e.g., \underline{A}C\underline{B}D\underline{A}E\underline{B}. In the resulting time series, the TSMD task is uniquely defined. To generate a time series with this structure that includes $\kappa'$ GT motif sets, at least $3\kappa'-1$ classes are required.\footnote{$\kappa'$ classes to construct the GT motif sets and $2\kappa'- 1$ classes to interleave the (at least) $2\kappa'$ instances of the GT motif sets with a single instance.} Hence, the maximum number of GT motif sets for a given classification dataset with $c$ classes is $\kappa'_{\max} = \lfloor \frac{c+1}{3} \rfloor$.

Our procedure generates $N$ time series of this structure from a z-normalized time series classification dataset with $c$ classes. For every generated time series, it samples $\kappa'$ uniformly from $[2:\kappa'_{\max}]$, and randomly selects the $\kappa'$ repeating classes.

\subsection{Setting 1: Variable-length motifs}
Our first experiment concerns the most general setting where occurrences of the same GT motif set can have different lengths. In this setting, we can only apply the methods that require $l_{\min}$ and $l_{\max}$ instead of the exact length $l$: LoCoMotif, Motiflets, and VALMOD (Section~\ref{sec:related_work}).\\

\noindent \textbf{Datasets.} We select suitable datasets from the variable-length classification datasets in the UCR and UEA Time Series Archive \citep{UCRArchive2018}. First, we select only datasets with more than five classes to be able to generate at least 2 GT motif sets per time series. Second, we select only datasets whose classes are distinguishable in an unsupervised manner. To this end, we cluster the datasets with the $k$-medoids algorithm using both DTW and Euclidean distance, and only select the datasets with an average Adjusted Rand Index (ARI) \citep{javed_benchmark_2020} greater than 0.5.\footnote{We z-normalize every instance before clustering. To apply Euclidean distance on a variable-length dataset, its instances are resampled to equalize their length to the average length.} Three datasets adhere to these criteria: \textsf{CharacterTrajectories},  \textsf{JapaneseVowels}, and \textsf{SpokenArabicDigits}. For each dataset, 20\% of the instances are used to generate 50 time series for hyperparameter tuning (constituting the validation set) and the remaining 80\% to generate 200 time series for evaluation, using the procedure described in Section~\ref{sec:generation}
 
 Note that the datasets we generate for this setting all resemble realistic TSMD problems. For example, each instance in the \textsf{SpokenArabicDigits} dataset represents a digit pronounced by a person. By concatenating multiple instances, we create a TSMD task that corresponds with finding the occurrences of every repeated digit in an audio recording; which is a task that could occur in a realistic scenario. \\

\begin{figure}
    \centering
    \includegraphics[width=\linewidth]{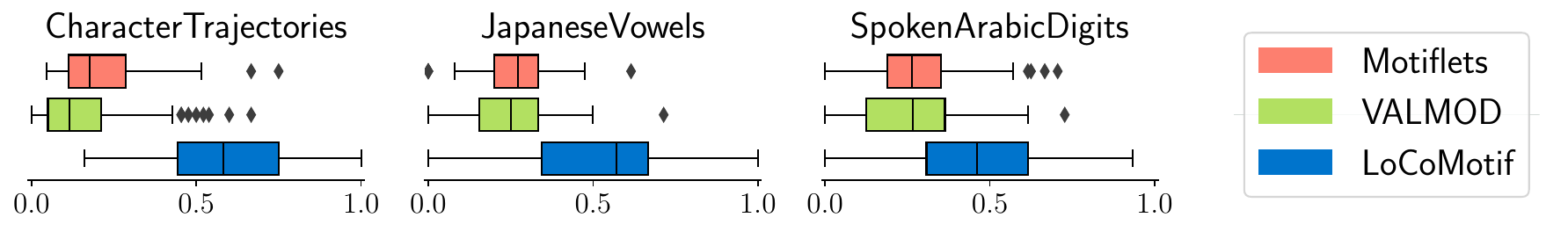}
    \caption{The distribution of the F1-score over the time series in every variable-length dataset. The three datasets are multivariate and resemble realistic problems. LoCoMotif achieves a higher average F1-score than its competitors.}
    \label{fig:boxplots_varlen}
\end{figure}
\begin{figure}
    \centering
    \includegraphics[width=0.5\linewidth]{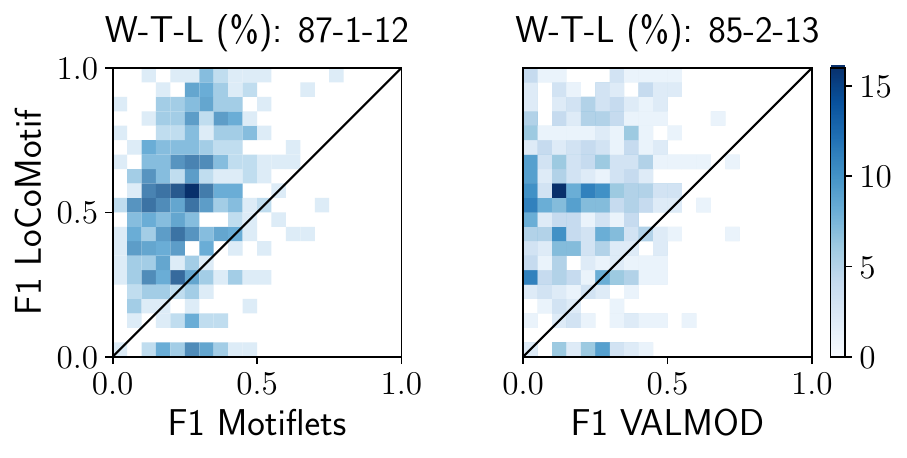}
    \caption{2D histograms that visualize the F1-score of LoCoMotif against those of Motiflets and VALMOD across all 600 time series. Both histograms show a higher concentration above the diagonal line, indicating that LoCoMotif outperforms its respective competitor in most cases. The percentage of \underline{W}ins, \underline{T}ies, and \underline{L}osses of LoCoMotif against every competitor are also shown.}
    \label{fig:histplots_varlen}
\end{figure}

\noindent \textbf{Hyperparameters.} For each method, we set $l_{\min}$ to the $0.1$-quantile and $l_{\max}$ to the $0.9$-quantile of the lengths of the instances in the validation dataset (to be insensitive to instances with unusually short and long lengths) and use $\kappa = \kappa'_{\max}$ (Section ~\ref{sec:generation}). The hyperparameter $k_{\max}$ that is required by Motiflets (Section~\ref{sec:related_work}) is set to $k_{\max}=2 k'_{\max}$ where $k'_{\max}$ is the actual value in the validation dataset.\footnote{Since Motiflets finds the best $k$ using the elbow method, it is better to specify $k_{\max}$ higher than the actual value.} We set the data-dependent hyperparameters required by LoCoMotif and VALMOD ($\rho$ and $r_f$, respectively) to maximize the average F1-score on the validation set.\footnote{For an unsupervised task, tuning the hyperparameters on a validation set is argued to be a sensible approach by \citet{soenen}.} For LoCoMotif, we consider $\rho \in \{0.3, 0.4, \dots, 0.9\}$ and a binary hyperparameter that determines whether to use time warping or not (Section~\ref{sec:variants}). For VALMOD, we use $r_{f} \in [2 : 6]$ as in the original paper \citep{linardi_matrix_2018}.
To apply the existing methods (that support only univariate time series) on a multivariate dataset, we select the dimension that leads to the highest F1-score on the entire validation set. \\

\noindent \textbf{Results.} LoCoMotif substantially outperforms its competitors in terms of F1-score for the three datasets (Fig.~\ref{fig:boxplots_varlen}). When combining the results, LoCoMotif achieves a win rate of $87\%$ against Motiflets and $85\%$ against VALMOD on all 600 generated time series (see Fig.~\ref{fig:histplots_varlen} for the distributions of the F1-score of LoCoMotif against its competitors). The high variability in the F1-score of every method can be explained by the varying difficulty among the time series, as some may contain GT motif sets constructed from classes that are inherently difficult to distinguish in an unsupervised manner. 

LoCoMotif uses time warping for the \textsf{CharacterTrajectories} and \textsf{SpokenArabicDigits} datasets (as a result of hyperparameter optimization). This is expected, given that the instances of the same class in these datasets exhibit temporal variations (such as the same character handwritten at different speeds).

\subsection{Setting 2: Fixed-length motifs}\label{sec:fixedlength}
Our second experiment investigates the specific setting where all motifs of all GT motif sets have a fixed length $l$. We further distinguish two cases: (1) a range around $l$ is known and (2) the exact value of $l$ is known. \\

\begin{figure}
    \centering
    \begin{subfigure}[b]{\linewidth}
        \includegraphics[width=\linewidth]{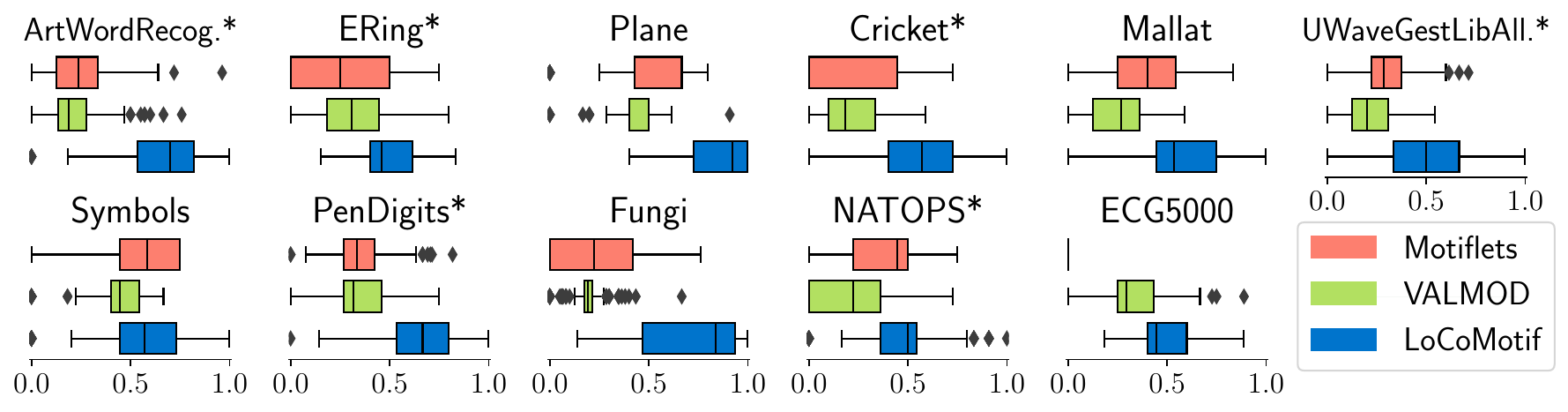}
        \caption{$\epsilon=0.25$}
        \label{fig:boxplots_epsilon=0.25.pdf}
    \end{subfigure}
    \begin{subfigure}[b]{\linewidth}
        \includegraphics[width=\linewidth]{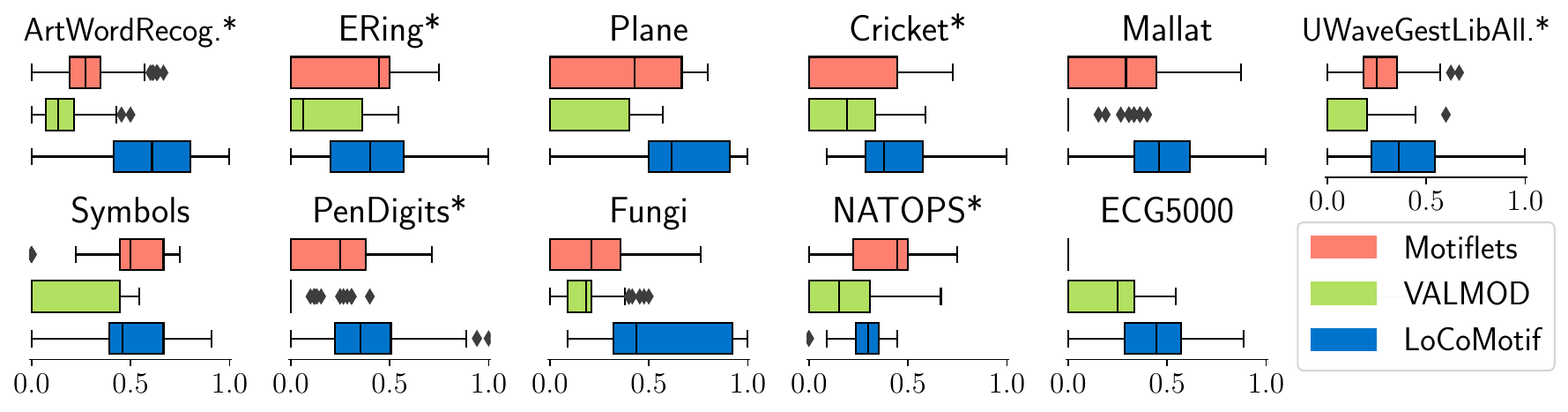}
        \caption{$\epsilon=0.50$}
        \label{fig:boxplots_epsilon=0.5.pdf}
    \end{subfigure}
    \caption{When all methods use a range around the exact length $l$ (determined by $\epsilon$), LoCoMotif achieves a higher average F1-score for all datasets, except for the \textsf{NATOPS} dataset when $\epsilon=0.5$. Multivariate datasets are annotated with *.}\label{fig:boxplots_epsilon}
\end{figure}

\noindent \textbf{Datasets.} For both cases, we use the fixed-length datasets from the UCR and UEA Time Series Archive \citep{UCRArchive2018}, again selecting datasets with more than five classes. In this case, we use a recent clustering benchmark paper \citep{javed_benchmark_2020} to select the datasets that have an average ARI of at least 0.5 averaged over multiple distance measures and over different clustering techniques. Five univariate and six multivariate datasets (listed in Fig.~\ref{fig:boxplots_epsilon}) adhere to these criteria. 

\begin{figure}
    \begin{subfigure}[b]{0.49\linewidth}
    \centering
    \includegraphics[width=\linewidth]{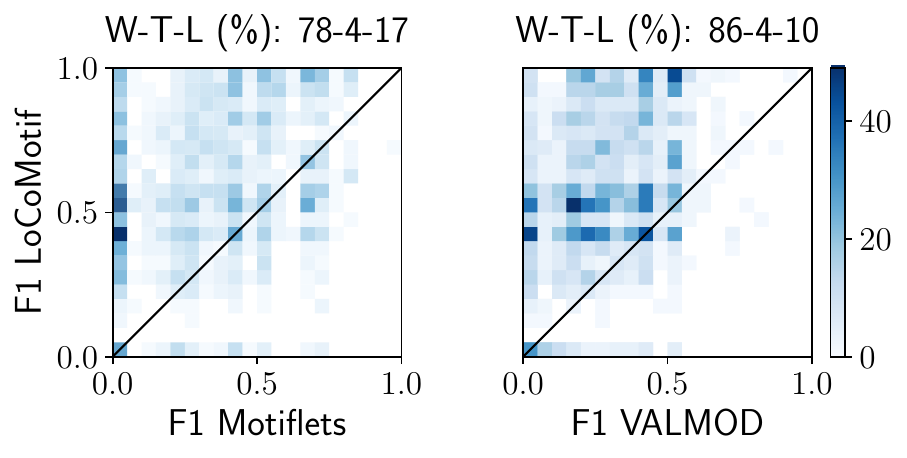}
    \caption{$\epsilon=0.25$}
    \label{fig:histplots_epsilon=0.25}
    \end{subfigure}
    \hfill
    \begin{subfigure}[b]{0.49\linewidth}
    \centering
    \includegraphics[width=\linewidth]{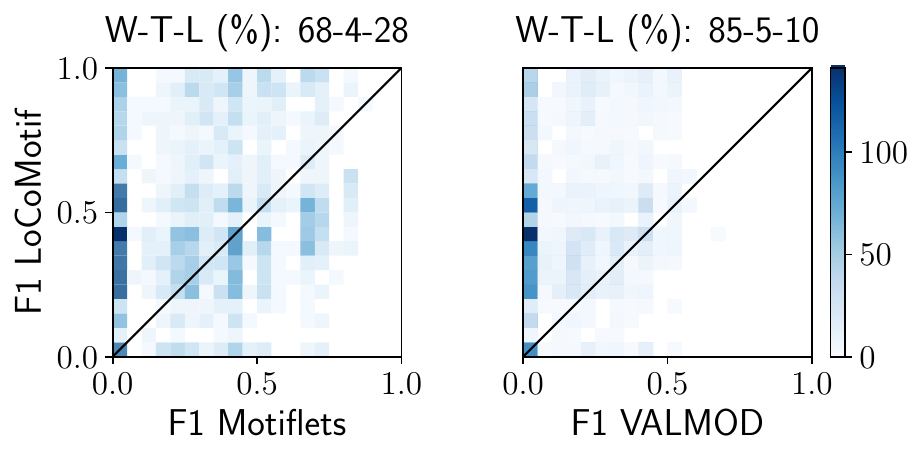}
    \caption{$\epsilon=0.50$}
    \label{fig:histplots_epsilon=0.50}
    \end{subfigure}
    \caption{When using a range (determined by $\epsilon$) around the exact motif length, LoCoMotif achieves a higher F1-score than Motiflets and VALMOD on the vast majority of all 2200 time series of the fixed-length datasets.}
    \label{fig:histplots_epsilon}
\end{figure}

\subsubsection{Range around $l$ is known}
To simulate the scenario where a user is uncertain about the exact length $l$ of the motifs in a time series, we use a length range around $l$. We compare the performance of the applicable methods (LoCoMotif, Motiflets, VALMOD) and examine their sensitivity to the user's uncertainty about motif length $l$ by gradually increasing the size of the length range.\\ 

\noindent \textbf{Hyperparameters.} We set $l_{\min} = \lfloor l-\epsilon l\rfloor$ and $l_{\max} = \lfloor l+\epsilon l\rfloor$ where $\epsilon \in \{0.25, 0.50\}$. The remaining hyperparameters are set in the same way as for Setting 1. \\

\noindent \textbf{Results.} For both length ranges, LoCoMotif achieves a higher F1-score than its competitors for the majority of the datasets (see Fig.~\ref{fig:boxplots_epsilon} for the distribution of the F1-score per dataset and Fig.~\ref{fig:histplots_epsilon} for the comparative distributio). When  
$\epsilon=0.25$, the win rates of LoCoMotif are 78\% against Motiflets and 86\% against VALMOD. When $\epsilon=0.50$, they are respectively equal to 68\% and 85\%. Motiflets seems to be the least sensitive to increasing $\epsilon$, which is expected because it first selects the most promising length within the specified range, and then finds fixed-length motif sets of that length (Section~\ref{sec:related_work}).

\subsubsection{$l$ is exactly known}\label{sec:fixedl}
In the most specific case where motifs have a fixed length $l$ that is known (equivalent to $\epsilon=0$), we can also include the fixed-length TSMD methods SetFinder and LatentMotifs in our comparative evaluation. \\

\begin{figure}
    \centering
    \includegraphics[width=\linewidth]{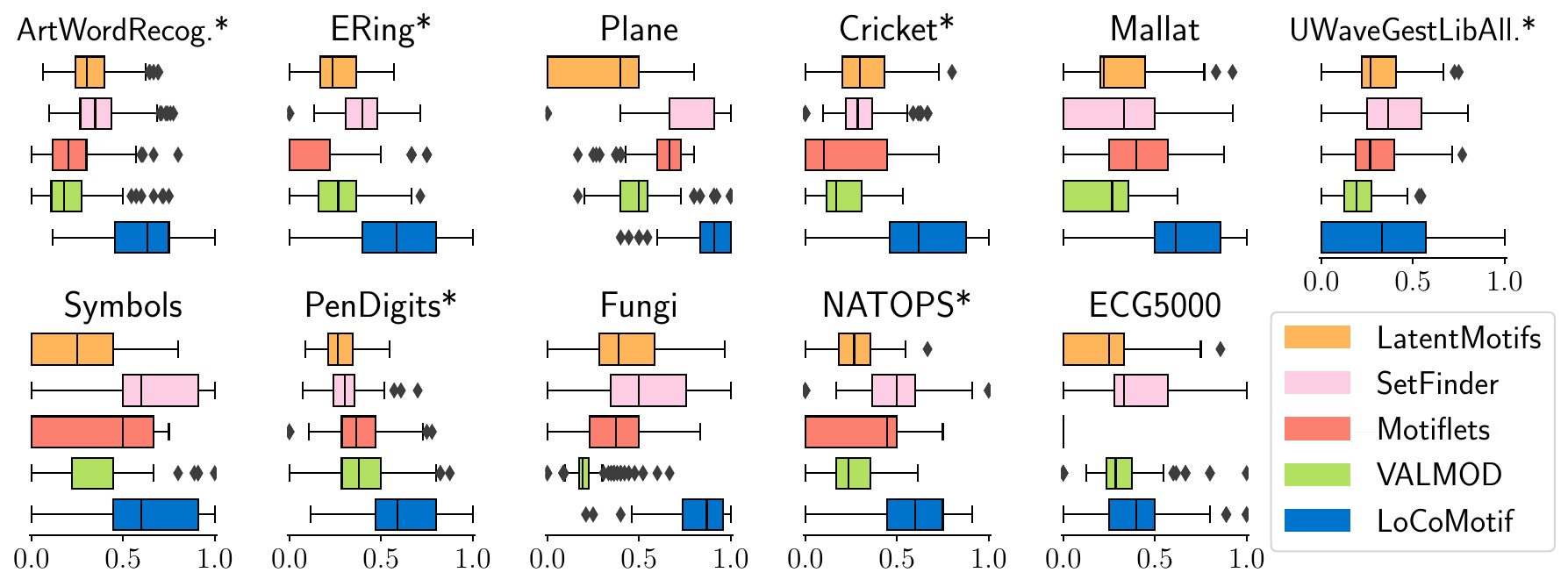}
    \caption{LoCoMotif achieves a higher average F1-score for all datasets except the \textsf{Symbols} dataset when all methods use the exact length $l$.}
    \label{fig:boxplots_epsilon=0.0}
\end{figure}
\begin{figure}
    \centering
    \includegraphics[width=\linewidth]{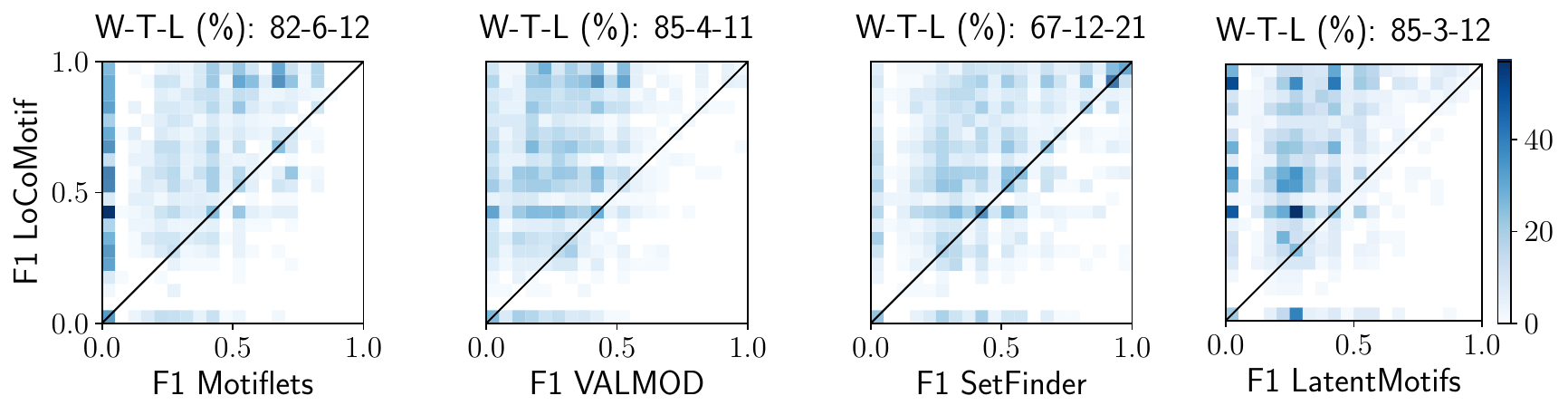}
    \caption{When all methods use the exact length $l$, LoCoMotif achieves a higher F1-score than its competitors for the  vast majority of the 2200 time series of the fixed-length datasets.}
    \label{fig:histplots_epsilon=0.0}
\end{figure}

\noindent \textbf{Hyperparameters.} All methods use the exact value of $l$ (methods that require a length range use $l_{\min} = l_{\max} = l$). For SetFinder and LatentMotifs, we specify hyperparameter $r$ (Section \ref{sec:related_work}) as correctly as possible using the validation dataset: for each class, we find the subsequence that minimizes the maximum distance (in terms of ZED) to all other subsequences in that class. Then, we set $r$ to the maximum distance over all classes. The rest of the hyperparameters are set as in Setting 1. \\

\noindent \textbf{Results.} In this case, LatentMotifs and SetFinder have an advantage because they are specifically designed to find fixed-length motif sets. Nevertheless, LoCoMotif outperforms all its competitors for the majority of the datasets (Fig.~\ref{fig:boxplots_epsilon=0.0}), achieving win rates equal to 82\%, 85\%, 67\%, and 85\% against Motiflets, VALMOD, SetFinder, and LatentMotifs, respectively (see Fig.~\ref{fig:histplots_epsilon=0.0}). 

LoCoMotif uses time warping for the \textsf{Cricket} and \textsf{ECG5000} datasets (determined based on the validation set). This is consistent with the fact that the first contains gestures (measured with multiple accelerometers) and the second contains heartbeats, both of which exhibit temporal variations.

\subsection{Runtimes}
\begin{figure}
    \centering
    \includegraphics[width=\linewidth]{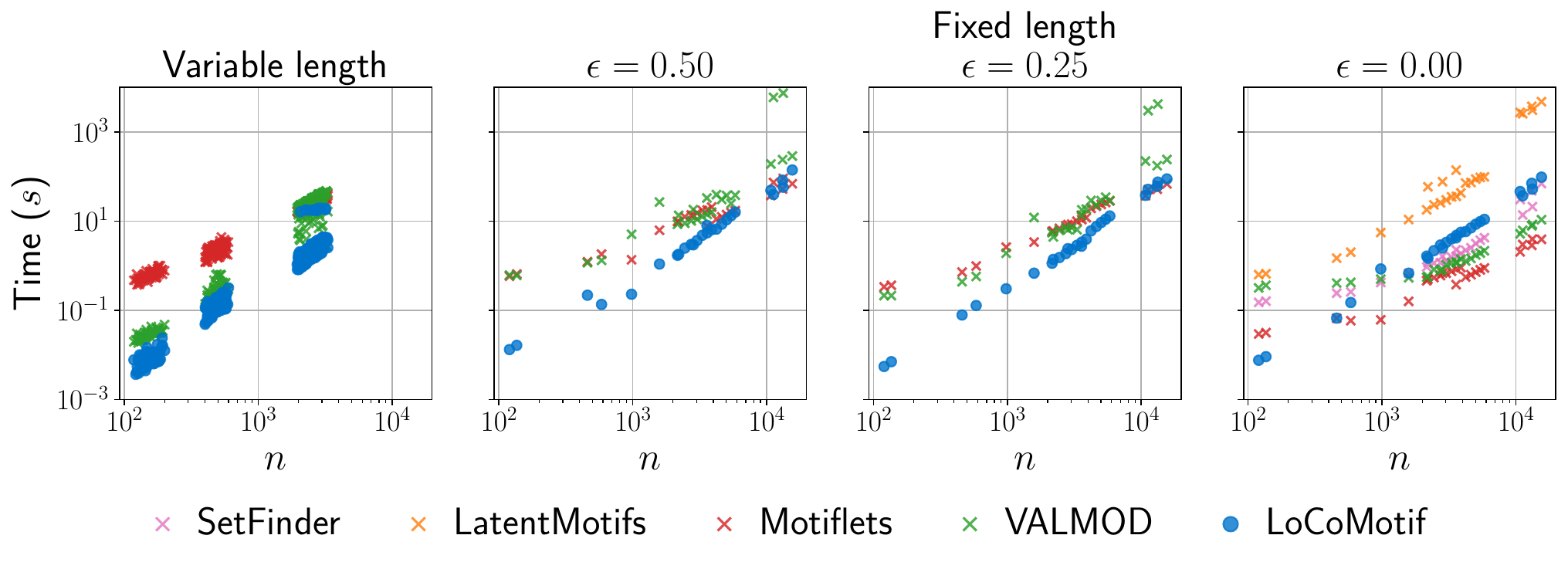}
    \caption{LoCoMotif is faster than existing methods in most of the cases.}
    \label{fig:runtimes}
\end{figure}

Lastly, we briefly compare the runtimes of LoCoMotif to those of existing methods (Fig.~ \ref{fig:runtimes}). When interpreting the results, two aspects should be kept in mind: the time series in our experiments have a relatively short length with respect to other studies, and the original implementations of the methods use different programming languages.\footnote{ LoCoMotif and Motiflets are implemented in Python, SetFinder and LatentMotifs in Java, and VALMOD in C.} All experiments are ran on the same computer as Section \ref{fig:results_physicaltherapy}.

LoCoMotif is faster than its competitors in the cases where a length range is used (variable length, and fixed length with $\epsilon \in \{ 0.25, 0.5 \}$). For a single length (that is, $\epsilon = 0$), LoCoMotif is slower than most existing methods, because the LoCo step has a computation time independent of the size of the length range and therefore determines the total computation time. However, LoCoMotif can still be preferred in this specific case, because it obtains the highest quality motif sets (as shown in Section \ref{sec:fixedl}).

\section{Conclusion and Future Work}\label{sec:conclusion}
We have presented LoCoMotif, a novel method for TSMD that offers unique capabilities among existing methods: it discovers multiple, variable-length motif sets in multivariate time series and uses time warping. Our proposed method provides these capabilities without compromising time complexity (except for the specific case where all motifs have equal length and it is known to the user).

We have shown the value of LoCoMotif in two ways. First, we applied it to a physiotherapy use case to which existing methods were not applicable. We concluded that LoCoMotif can accurately detect the exercise executions in a sensor-recorded physiotherapy session, and separate them based on their correctness in an unsupervised manner. Second, we comparatively evaluated LoCoMotif against existing methods using benchmark datasets. LoCoMotif outperformed its competitors in multiple settings, including settings for which the existing methods are specialized.

Future work could extend LoCoMotif to operate in an active-learning setting. In this case, LoCoMotif would interact with the user to obtain the motifs they prefer. \\

\noindent \textbf{Acknowledgements} This research received funding from the Flemish Government under the ``Onderzoeksprogramma Artificiële Intelligentie (AI) Vlaanderen'' programme and the VLAIO ICON-AI Conscious project (HBC.2020.2795).

\bibliography{references}

\end{document}